# Performance, Successes and Limitations of Deep Learning Semantic Segmentation of Multiple Defects in Transmission Electron Micrographs

Ryan Jacobs[1,*], Mingren Shen[1,*], Yuhan Liu[2,*], Wei Hao[2], Xiaoshan Li[2], Ruoyu He[2], Jacob RC Greaves[1], Donglin Wang[2], Zeming Xie[2], Zitong Huang[3], Chao Wang[2], Kevin G. Field[4], Dane Morgan[1]

[1]Department of Materials Science and Engineering, University of Wisconsin-Madison, Madison, Wisconsin, 53706, USA

[2]Department of Computer Sciences, University of Wisconsin–Madison, Madison, Wisconsin, 53706, USA

[3]Department of Electrical and Computer Engineering, University of Wisconsin–Madison, Madison, Wisconsin, 53706, USA

[4]Nuclear Engineering and Radiological Sciences, University of Michigan - Ann Arbor, Michigan, 48109 USA

*These authors contributed equally

## Abstract

In this work, we perform semantic segmentation of multiple defect types in electron microscopy images of irradiated FeCrAl alloys using a deep learning Mask Regional Convolutional Neural Network (Mask R-CNN) model. We conduct an in-depth analysis of key model performance statistics, with a focus on quantities such as predicted distributions of defect shapes, defect sizes, and defect areal densities relevant to informing modeling and understanding of irradiated Fe-based materials properties. To better understand the performance and present limitations of the model, we provide examples of useful evaluation tests which include a suite of random splits, and dataset size-dependent and domain-targeted cross validation tests. Overall, we find that the current model is a fast, effective tool for automatically characterizing and quantifying multiple defect types in microscopy images, with a level of accuracy on par with human domain expert labelers. More specifically, the model can achieve average defect identification F1 scores as high as 0.8, and, based on random cross validation, have low overall average (+/- standard deviation) defect size and density percentage errors of 7.3 (+/- 3.8)% and 12.7 (+/- 5.3)%, respectively. Further, our model predicts the expected material hardening to within 10-20 MPa (about 10% of total hardening), which is about the same error level as experiments. Our targeted evaluation tests also suggest the best path toward improving future models is not expanding existing databases with more labeled images but instead data additions that target weak points of the model domain, such as images from different microscopes, imaging conditions, irradiation environments, and alloy types. Potential limitations of the present model based on practical use cases such as extending predictions on new images outside the domain of the training data. Finally, we discuss the first phase of an effort to provide an easy-to-use, open-source object detection tool to the broader community for identifying defects in new images.

## Introduction

Extended defects in materials are critical in determining their properties and performance. The role of defects are particularly important for materials performance in extreme environments, where a cornerstone of advanced materials discovery and development is the understanding of the production and evolution of defects. In many cases, extreme environments include elevated temperatures, stress, corrosion rates, and radiation, which can lead to the production of defects including point defects, line dislocations, dislocation loops, cavities/voids, stacking fault tetrahedra, and precipitates, to name a few. The nucleation, growth and evolution of these various defect types can lead to deleterious changes in materials performance, including the loss of strength and ductility. Common, simplified structure-property relationships such as the dispersed barrier hardening model[1] show that these changes in properties are directly related to the size, number density and type of defects present. As a result, a significant portion of the materials discovery for extreme environments, development and deployment cycle is spent characterizing and quantifying these defects after simulated exposures. This characterization and quantification of defects is critical to predict and understand material performance in an array of complex and aggressive environments.

Transmission electron microscopy (TEM) is the method of choice for characterizing and quantifying defects in materials. Analyzing digitized TEM images is commonly done with software packages like ImageJ,[2] which enable a user to manually quantify the size, shape, and locations of defects in the images. This purely manual, human-based task is very time consuming, error prone, inconsistent, generally requires many hours of training and expertise to do well, and is not scalable to large dataset sizes. The latter point is particularly important, considering that modern TEM instruments can now routinely collect tens of thousands of images or hours of video content, the manual analysis of which is not feasible.[3] Therefore, the development of automated methods for quantifying and analyzing defects in TEM images, as well as understanding the advantages, shortcomings, and potential pitfalls of these methods can be used to establish a set of best practices for the community as these automated methods witness increased adoption.

The introduction of deep learning methods in 2012 revolutionized the field of computer vision,[4,5] and the maturation of these methods has direct implications for the present problem of automatically characterizing and quantifying defects in TEM images. Deep learning techniques typically involve the use of convolutional neural networks (CNNs) and have enabled stunning advances ranging from superhuman facial recognition to self-driving vehicles. As a prime example, yearly object classification competitions such as the Pattern Analysis, Statistical Modeling and Computational Learning Visual Object Classes (PASCAL VOC)[6] and ImageNet[7] Large Scale Visual Recognition Challenge (ILSVRC)[8] witnessed a significant advance in prediction accuracy after 2012 when the first deep learning-based image classification network, AlexNet,[5] enabled a performance increase from about 40% correct in the prior two years to nearly 60% correct in the PASCAL VOC challenge.[9] In the following few years, the advances in the deep learning object classification methods made these models so adept at classifying the test set images at these competitions, that as of 2018 the average classification performance was at or above 90% for PASCAL VOC, and greater than 80% for ILSVRC.[9]

The coupled use of traditional computer vision and machine learning methods, such as a workflow incorporating a sequence of blurring, thresholding, and masking operations combined with clustering algorithms or random forest classification models have yielded numerous successes in analyzing and quantifying an assortment of features in microscopy images.[10–12] However, traditional computer vision methods tend to suffer from reliance on empirically chosen parameters, making them useful for limited sets of cases and thus less general and less transferable than deep learning-based methods. Deep learning methods are increasingly being adopted in materials science.[13–17] In microstructure characterization in materials science,[18–20] the advances of these deep learning methods has enabled a shift from the combined use of manually implemented and tuned traditional computer vision and machine learning techniques to more automatic deep learning methods. The use of deep learning methods has shown success in tasks ranging from highlighting defective regions of crystalline materials in high resolution scanning TEM (STEM) images,[21] segmenting different microstructural phases,[22] finding locations of individual atoms in a material,[23] counting and analyzing nanoparticles,[24,25] identifying and classifying surface defect types in steels[26,27] and classifying types of dislocation loops at the microscale.[28,29]

In the past few years, there have been a handful of pioneering studies employing deep learning methods to characterize and quantify defects in electron microscopy images. The work of Li et al. used a standard CNN architecture coupled with traditional computer vision methods to quantify defects in FeCrAl alloys.[28] Two key limitations to the work from Li et al. were the ability to only identify a single type of defect, and the lack of pixel-level segmentation information from the model, prompting the use of traditional computer vision methods that required extensive manual tuning to obtain the desired performance. The study of Shen et al. extended the work of Li et al. by using the Faster R-CNN (regional convolutional neural network) algorithm on the same data from Li et al. and was able to characterize multiple defect types with a fully deep learning approach. However, this work still used traditional computer vision methods to extract details of predicted defect size.[29] In a similar vein, the work of Anderson et al. also used the Faster R-CNN algorithm to detect He bubbles, which are sometimes called cavities or voids, in irradiated Ni-based alloys. Like the works of Li et al. and Shen et al., this study also used additional post-processing methods separate from the deep learning model to extract materials property information such as void sizes, because the Faster R-CNN model does not provide pixel-level segmentation information.[30] In addition, Shen et al. also employed the YOLO (You Only Look Once) object detection model to demonstrate real-time identification and tracking of defect loops in FeCrAl alloys for sets of TEM images extracted from video.[31] As a final example, the work of Roberts et al. employed a model called DefectSegNet, based on the U-net model architecture, as the first study to demonstrate pixel-level segmentation of multiple defect types in electron microscopy images. This work, while very encouraging, does not conclusively demonstrate widespread effectiveness of pixel-wise segmentation models for two reasons. First, images were gathered for only a single material alloy and single sample, and two large 2048×2048 images were used for each defect type, which, after augmentation, amounted to 48 individual smaller training images, likely indicating a narrow model domain and small amount of training data. Second, the output of U-net models consists of a single mask for the entire image, denoting whether individual pixels are part of a defect or part of the background, thus making quantification of per-defect statistics such as size, shape, and density, more difficult, necessitating the use of additional techniques beyond the deep learning approach used for detection.[32]

In this study, we employ pixel-level segmentation models to create an automated, fully deep-learning based approach to classify and analyze multiple defect types in irradiated FeCrAl alloys. We highlight analysis of key model performance statistics, with a focus on quantities such as predicted distributions of defect shapes, defect sizes, and defect areal densities relevant to informing modeling and understanding of irradiated alloy materials properties. In addition, to better understand the performance and present limitations of the model, we provide examples of useful evaluation tests which include a suite of random splits, and dataset size-dependent and domain-targeted cross validation tests. Finally, a significant expansion of the labeling in the image database from the works of Li et al.[28] and Shen et al.[29] to include both more labeled images and to include pixel-level segmentation enables us to make a current best-fit segmentation model for identifying defect loops in irradiated FeCrAl alloys, which can be used by other researchers to make predictions of defects in new images. We have provided the final model fit to all images in the latest database and a Google Colab notebook to allow users to easily make predictions on new test images. This automated analysis provides output of numbers and locations of each defect and the test images with the predictions overlayed (see **Data Availability**).

## Data and Methods

The image database used in this study consists of FeCrAl alloys which have undergone neutron or ion irradiation. The images are exactly those available derived from a series of published studies from Field et al.[33–35], although some of the data has yet to be summarized in a publication and we have extended the labeling, as discussed below. The samples are all FeCrAl alloys but vary in composition, microstructure (including grain size and line dislocation density) and irradiation conditions. All images are from a single TEM imaging condition, specifically [100] on-zone bright field STEM. These imaging conditions produce defects appearing as black contrast features on a white background. In the case of irradiated FeCrAl, on-[100] zone imaging results in open single edge elliptical loops that are dislocation loops with a Burgers vector of $a_0/2 \langle 111 \rangle$ (henceforth referred to as ⟨111⟩ loops), open double edge elliptical loops and closed elliptical solid loops that are dislocation loops with a Burgers vector of $a_0\langle 100\rangle$ (henceforth referred to as ⟨100⟩ loops), and closed circular solid dots that are typically called black dot defects with a Burgers vector of either $a_0/2\langle 111\rangle$ or $a_0\langle 100\rangle$ (henceforth referred to as black dots). An example experimental micrograph showing the visual characteristics of each labeled defect type is shown in **Figure 1**.

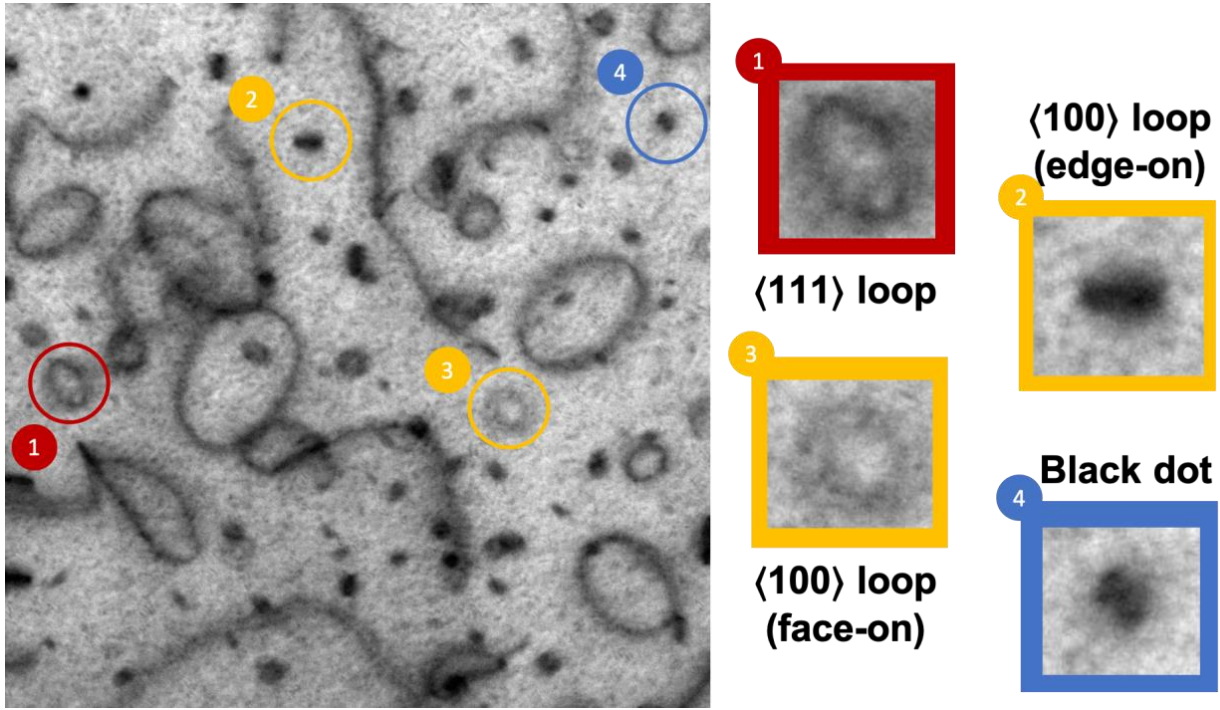

**Figure 1.** Example experimental image showing examples of the different irradiation-induced defect types characterized in this work. The red, yellow, and blue colors denote ⟨111⟩ loops, ⟨100⟩ loops, and black dot defects, respectively.

As mentioned above, the image database used in this work was previously used in the works of Li et al. and Shen et al., however these studies did not include pixel-level segmentation information. For this study, the image database was updated to include new labeling, specifically new ground truth pixel-level segmentation annotations. We developed three datasets of labels. The first considered a set of 107 images that were labeled with pixel-level segmentation by a first group of domain experts who found 5,382 defect instances. Note that this is not all the images in the full set of images. We call this set of 107 images and 5,382 defect instances "Dataset1". Then, to better understand how the labeling might impact results the same 107 images were labeled by a second set of domain experts, this time finding 5,053 defect instances. We call this set of 107 images and 5,053 labels "Dataset2". Finally, to explore how using a larger set of labeled images might impact the results, we labeled additional images and joined them with Dataset2. This led to a new dataset with 182 annotated images and 13,675 defect instances, which we denote as "Dataset2 expanded". **Table 1** contains a summary of the basic characteristics of each dataset, including number of images and number of each labeled defect type. Numerous different splits of train and test images and their associated defects are used throughout this work. **Table 2** provides a summary of the number of images and each defect type present in the various train and test datasets analyzed in this study. All segmentation mask annotations for both image datasets were made using the VGG Image Annotator web application.[36] All of the data for these three datasets has been made available on Figshare (see **Data Availability** section).

**Table 1.** Summary of labeled image dataset characteristics.

| Dataset | Number of Images | Number of $\langle 100 \rangle$ loops | Number of $\langle 111 \rangle$ loops | Number of black dots | Total number of defects |
|---|---|---|---|---|---|
| Dataset1 | 107 | 715 | 2353 | 2314 | 5382 |
| Dataset2 | 107 | 754 | 2395 | 1904 | 5053 |
| Dataset2 expanded | 182 | 2835 | 4602 | 6238 | 13,675 |

**Table 2.** Summary of number of images and defects for each dataset split used in this work. Test set values given in parentheses.

| Parent Dataset | Split Name | Number of train (test) images | Number of $\langle 100 \rangle$ loops for train (test) | Number of $\langle 111 \rangle$ loops for train (test) | Number of black dots for train (test) |
|---|---|---|---|---|---|
| Dataset1 | Initial split | 85 (22) | 587 (128) | 1909 (444) | 1835 (479) |
| Dataset1 | Random CV 1 | 86 (21) | 607 (108) | 1987 (366) | 1896 (418) |
| Dataset1 | Random CV 2 | 86 (21) | 566 (149) | 1802 (551) | 1878 (436) |
| Dataset1 | Random CV 3 | 86 (21) | 552 (163) | 1870 (483) | 1676 (638) |
| Dataset1 | Random CV 4 | 86 (21) | 597 (118) | 1926 (427) | 1887 (427) |
| Dataset2 | Initial split | 85 (22) | 600 (134) | 1955 (440) | 1548 (356) |
| Dataset2 expanded | Initial split | 160 (22) | 2701 (134) | 4162 (440) | 5882 (356) |
| Dataset2 expanded | Leave out 10%, run 1 | 164 (18) | 2423 (412) | 4146 (456) | 5468 (770) |
| Dataset2 expanded | Leave out 10%, run 2 | 164 (18) | 2617 (218) | 4160 (442) | 5800 (438) |
| Dataset2 expanded | Leave out 10%, run 3 | 164 (18) | 2500 (335) | 4180 (422) | 5280 (958) |
| Dataset2 expanded | Leave out 25%, run 1 | 137 (45) | 1863 (972) | 3342 (1260) | 4341 (1897) |

| Dataset2 expanded | Leave out 25%, run 2 | 137 (45) | 2329 (506) | 3553 (1049) | 5031 (1207) |
| Dataset2 expanded | Leave out 25%, run 3 | 137 (45) | 2179 (656) | 3454 (1148) | 4504 (1734) |
| Dataset2 expanded | Leave out 50%, run 1 | 91 (91) | 1270 (1565) | 2290 (2312) | 2884 (3354) |
| Dataset2 expanded | Leave out 50%, run 2 | 91 (91) | 1635 (1200) | 2321 (2281) | 3552 (2686) |
| Dataset2 expanded | Leave out 50%, run 3 | 91 (91) | 1657 (1178) | 2336 (2266) | 3303 (2935) |
| Dataset2 expanded | Leave out 75%, run 1 | 46 (136) | 646 (2189) | 1179 (3423) | 1589 (4649) |
| Dataset2 expanded | Leave out 75%, run 2 | 46 (136) | 846 (1989) | 1215 (3387) | 1704 (4534) |
| Dataset2 expanded | Leave out 75%, run 3 | 46 (136) | 778 (2057) | 1243 (3359) | 1459 (4779) |
| Dataset2 expanded | Leave out 90%, run 1 | 18 (164) | 219 (2616) | 454 (4148) | 618 (5620) |
| Dataset2 expanded | Leave out 90%, run 2 | 18 (164) | 175 (2660) | 513 (4089) | 497 (5741) |
| Dataset2 expanded | Leave out 90%, run 3 | 18 (164) | 329 (2506) | 459 (4143) | 622 (5616) |
| Dataset2 | Leave out irradiation | 12 (9) | 58 (47) | 195 (334) | 117 (268) |
| Dataset2 | Leave out alloy | 9 (51) | 47 (1724) | 334 (1842) | 268 (3271) |
| Dataset2 | Leave out microscope/sample | 18 (70) | 216 (492) | 792 (1424) | 598 (1369) |
| Dataset2 expanded | Leave out irradiation | 21 (9) | 210 (47) | 423 (334) | 707 (268) |
| Dataset2 expanded | Leave out alloy | 18 (51) | 314 (1724) | 651 (1842) | 767 (3271) |
| Dataset2 expanded | Leave out microscope/sample | 69 (70) | 2038 (492) | 2493 (1424) | 4038 (1369) |

Throughout this study, we use the Mask R-CNN model as implemented in the Detectron2 package, which uses PyTorch as the backend. The Detectron2 package was developed by the Facebook AI Research (FAIR) team.[37] Detectron2 is freely available and enables implementation of many object detection models, such as Faster R-CNN,[38] Mask R-CNN,[39] and Cascade R-CNN.[40] These object detection models have been pre-trained on either the ImageNet[7] or Microsoft COCO[41] (Common Objects in Context) image databases, enabling use of the transfer learning technique. When using transfer learning, the model backbone weights are frozen to those obtained from the previous ImageNet or Microsoft COCO image training, save for a small number of terminal layers (2 throughout this work). The weights in these terminal layers are then updated during the training process to tune the model for the particular application of interest, in this case detecting certain defect types in electron microscopy images. All post-processing of Mask R-CNN model predictions and associated analysis was performed using in-house Python scripts, which we have made available on Figshare (see **Data Availability** section).

In this work, we evaluate the performance of our Mask R-CNN models on a number of different application-specific test central to understanding the impact of different defect types on the mechanical properties of an irradiated alloy. These test include how well the model can predict the areal density and size of defects in an image, and how well the model can discern the location and type of defects in an image. Explanations of the key we quantify to evaluate the overall performance of the Mask R-CNN model are summarized in **Table 3**. Note that the Heywood circularity factor is defined as the perimeter of an object divided by the circumference of a circle of the same area.

When training and using object detection models, a key performance parameter to choose is that of the intersection-over-union (IoU) score. The IoU score is used as a threshold value to decide whether a predicted object mask overlaps sufficiently with a ground truth mask such that the prediction can be considered a successfully "found" object. When evaluating an image, there is a list of true defect masks and predicted defect masks. To decide whether a defect has been found in the correct location, the IoU of every predicted defect is calculated for each true defect, and the defect with the highest IoU score is considered the best possible match. Then, if this computed IoU score is above the designated threshold, this predicted defect is considered to be found. Each true defect can only be found one time, so if multiple predicted defects are found to pass the IoU threshold with a particular true defect, the predicted defect with the highest IoU score is considered the found defect, and the other defect(s) would then be considered false positives.

**Table 3.** Summary of key model test statistics used in this work.

| Model test statistic | Explanation of statistic |
|---|---|
| Defect find test | Precision (P), recall (R), and F1 score of whether defect was found at correct position within some intersection-over-union (IoU) tolerance, regardless if defect is correct type |
| Defect type test | P, R, and F1 score of defect type when defect was found at correct position within some IoU tolerance |
| Defect areal density | Percent error between true and predicted defect areal densities (i.e. number of each defect type per square nanometer) |
| Defect size | Percent error between true and predicted defect sizes |
| Defect shape | Percent error between true and predicted defect shapes using the Heywood circularity factor |

In addition to the particular set of application-specific test statistics summarized in **Table 3**, we performed a number of different detailed test types. A summary of the different types of tests performed, what aspects of the model or data are changed in each test, and the rationale for performing each test is provided in **Table 4**. These different test types, particularly assessing the impact of different train/test image splits, dataset size, and impact of ground truth labels, may serve as a basis for better understanding the successes and limitations of object detection models, especially in the context of characterizing and quantifying objects in electron microscopy images.

**Table 4.** Summary of each basic test performed, and what can be learned by performing each test.

| Test type | What is changed in model, data, or analysis? | What can be learned from the test? |
|---|---|---|
| Effect of train/test split: random cross validation | Images and their associated annotation files used to train the model and subsequently used in model testing are shuffled randomly | How choice of which images are used to train the model and which are used to test the model may impact the model performance. Random shuffling ensures the train and test sets belong to approximately the same domain |
| Effect of train/test split: targeted group cross validation | Images and their associated annotation files used to train the model and subsequently used in model testing are split into physically motivated groups, e.g., leave out alloy, leave out irradiation condition, etc. | How choice of which images are used to train the model and which are used to test the model may impact the model performance. Grouped cross validation means the train and test sets may be in different domains, providing an assessment of model domain of applicability |
| Effect of data ground truth labeling | Which ground truth dataset (images and their annotations) are used, i.e. whether Dataset1 or Dataset2 is used for training and testing | Impact of ground truth labeling on resulting model performance |
| Effect of data set size | Size of training set used | Impact of training dataset size on resulting model performance |

# Results and Discussion

**Assessing performance of model on single dataset**

In this section, we assess the Mask R-CNN model performance using the best set of hyperparameters obtained from a preliminary survey of roughly 25 Mask R-CNN model runs (see **Supporting Information**). All fits in this section are performed on a single dataset, Dataset1 "initial split". **Figure 2** provides a graphical representation of the calculated precision, recall, and F1 score for the test of finding defects (regardless of whether type is correct) as a function of this IoU cutoff. We have found an IoU=0.3 provides a reliable balance of model performance for this defect find test while also providing reliable predictions of defect sizes, shapes and densities (to be discussed later). In **Figure 2**, the Mask R-CNN overall F1 score at IoU=0.3 is about 0.8, which is nearly identical to the value obtained from Shen et al., who used the Faster R-CNN model as implemented in the ChainerCV package.[29,42] This result indicates that the Mask R-CNN model used in this work can provide defect find statistics at the same level of quality as Faster R-CNN, and that the use of Detectron2 vs. ChainerCV and different backbone structure (ResNet 50 here, VGG16 in Shen et al.) does not appreciably alter the model quality, at least for this case. **Figure 3** provides three sets of images, comparing the ground truth labels with the Mask R-CNN model predictions. Similar to what was observed in the work of Shen et al., from manual inspection the object detection model does well overall at correctly categorizing and placing defect locations on the image relative to the ground truth. There are some observable errors in the prediction vs. the ground truth, such as missing some defects which should be present (false negative), predicting some defects to be present which should not be (false positive), and mis-categorizing some defects. These types of errors are all to be expected, and more details on their discussion and quantification were provided in the study of Shen et al.[29]

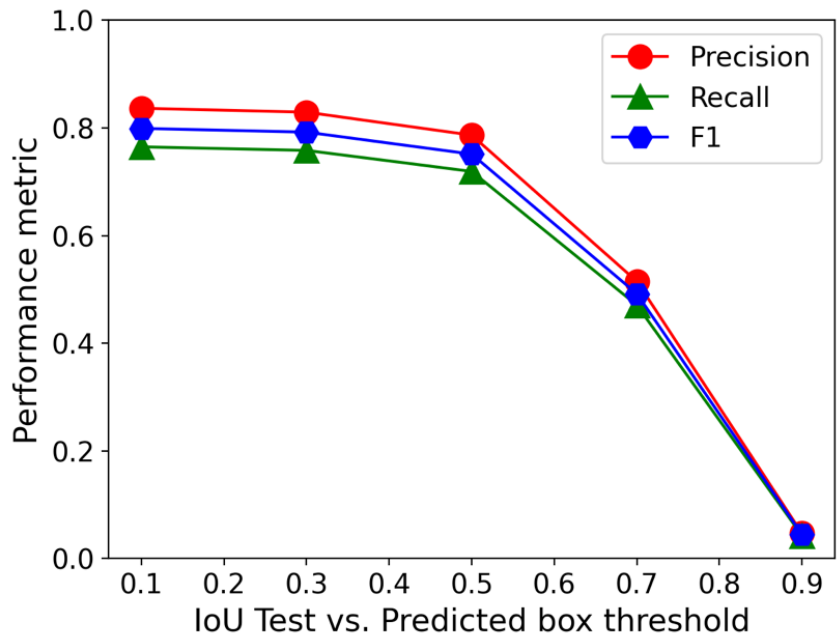

**Figure 2.** Model performance as a function of IoU cutoff between predicted and ground truth. The model was fit and evaluated using Dataset1 "initial split".

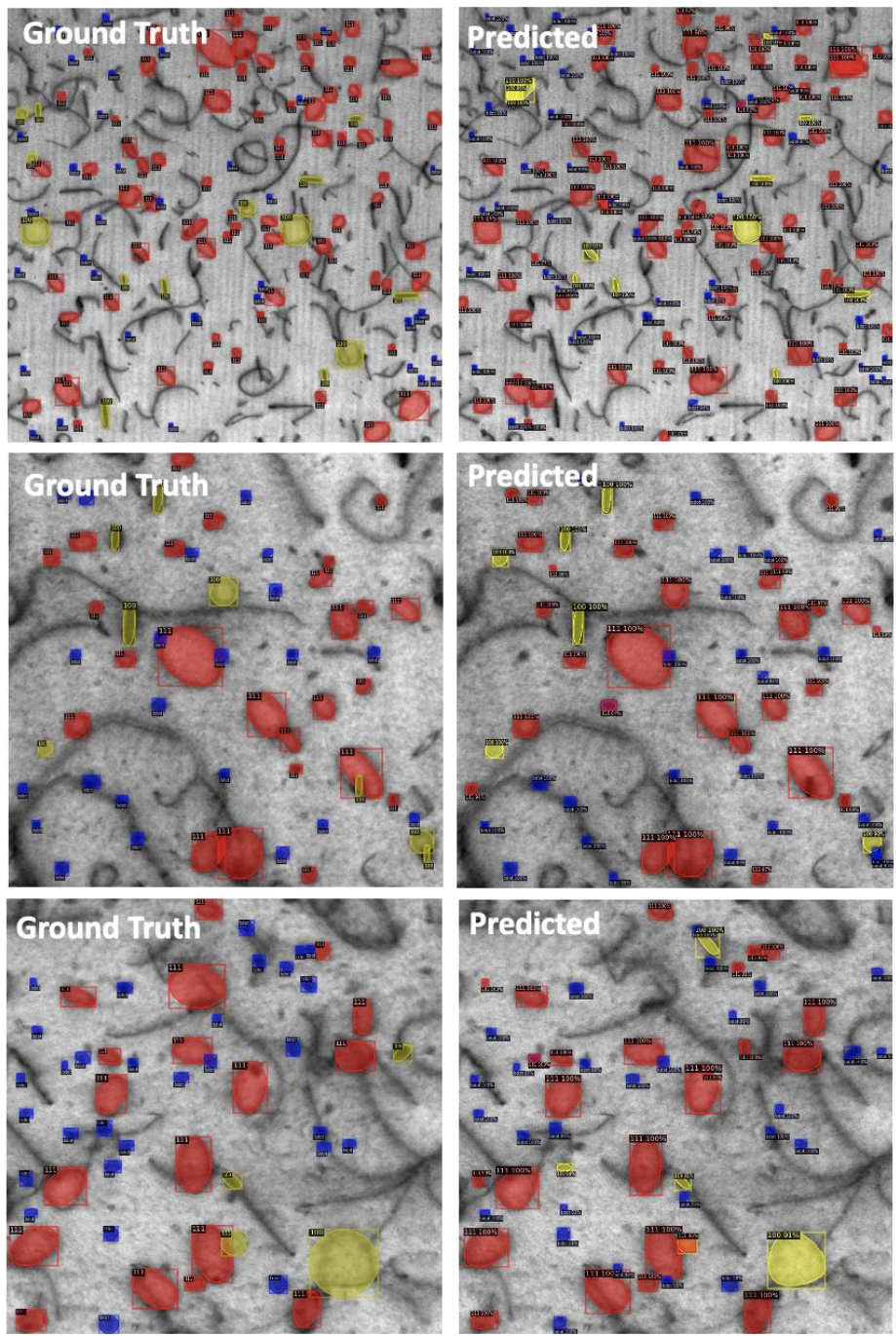

**Figure 3.** Examples of labeled ground truth (left columns) and Mask R-CNN predicted (right column) images. The red, yellow, and blue masks denote ⟨111⟩ loops, ⟨100⟩ loops and black dot defects, respectively. The predictions shown here were made with IoU=0.3 from a model fit and evaluated on Dataset1 "initial split".

**Detailed materials-centric property statistics obtainable from Mask R-CNN model**

In this section, we present a discussion of materials-centric properties obtained from the Mask R-CNN model predictions, specifically the distributions of predicted vs. true defect sizes, shapes, areal densities, and an approximation of the expected increase in yield stress based on a dispersion hardening model. Throughout this section, fits to Dataset1 "initial split" are used, and an IoU value of 0.3 is used based on the discussion in the previous section. **Figure 4** shows histogram distributions of true and predicted values of defect shape and defect size. We examine two cases for each distribution: the case where all true and predicted defects are used in the analysis, and a second case examining only the instances where a defect was found in the correct location, based on the implemented IoU=0.3. These two situations provide us with slightly different information regarding the model performance. For the situation assessing all defects (**Figure 4A** and **Figure 4C**), this comparison is indicative of the errors one may expect for applying the model to new test images where the number and locations of defects are not known *a priori*. For the situation assessing only found defects (**Figure 4B** and **Figure 4D**), this comparison is indicative of how well the model can predict the size and shape of defects for the case where it has explicitly found a defect in the correct location. From **Figure 4**, the error in the mean values of defect shape and size when considering all true and predicted defects are nearly 0% (accurate to two decimal places) and 7.1%, respectively. Qualitatively, in **Figure 4** the distributions between true and predicted defects are generally in very good agreement, and the distributions match more closely for the case of comparing found defects only. This result makes sense, given that comparing the distributions between all true and predicted defects will have contributions from false positives and false negatives which is expected to alter the overall distribution compared to only comparing correctly identified defects. Note that the large fractional errors observed for larger values of Heywood circularity above about 1.5 are for bins with < 10 defects and therefore sensitive to small counting errors. Also, defects with Heywood circularity above about 1.5 almost always consist of long edge-on ⟨100⟩ loops, which we speculate the model undercounts as they may be confused with pre-existing line dislocations, which are considered part of the image background and not a feature of interest.

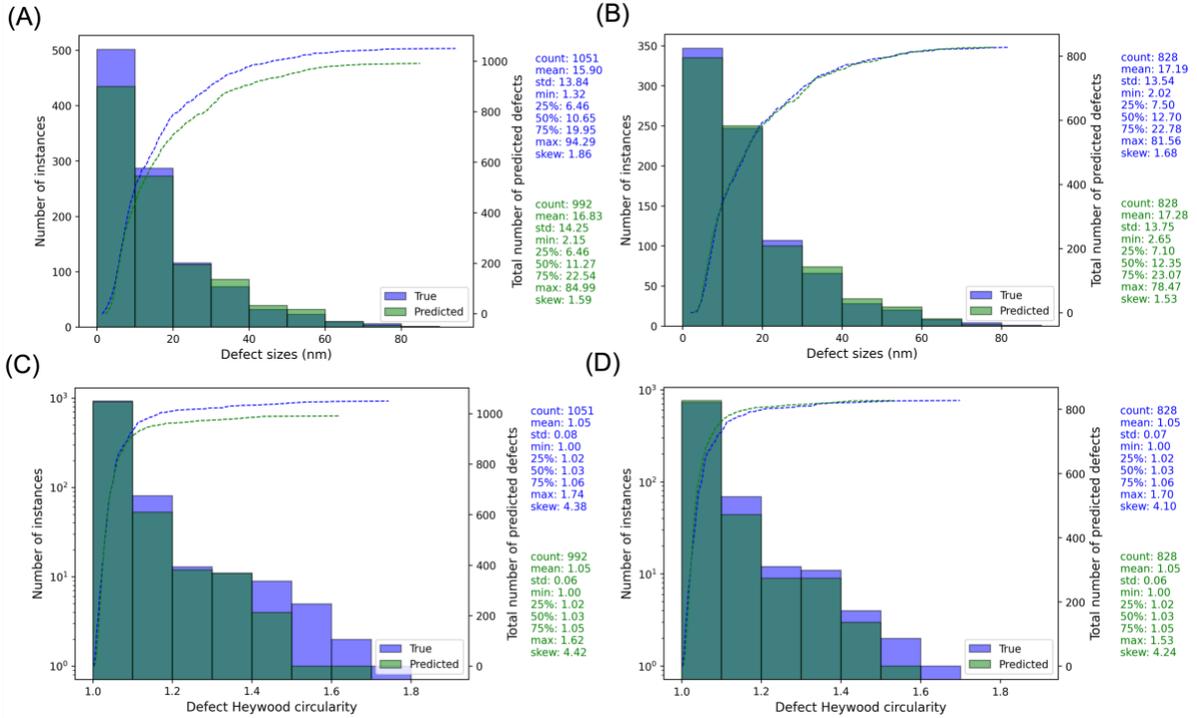

**Figure 4.** Histograms comparing distributions of true and predicted defect sizes (A, B) and defect shapes (C, D), computed as the Heywood circularity, for all true and predicted defects (A, C) and only those defects found in the correct location (IoU = 0.3) (B, D). Note that the defect number histograms in (C, D) are log scale.

In **Figure 5**, we take the defect size distribution data for all true and predicted defects from **Figure 4A** and break it up to be on a per-defect type basis. In **Figure 5**, the shapes of the predicted defect size distributions match well with the true distributions, though two deviations are notable. First, in **Figure 5A** the predicted black dot size distribution skews toward values smaller on average than the true values. Second, in **Figure 5B** the number of predicted instances of ⟨111⟩ loops are slightly overestimated in their number and in **Figure 5C** the instances of ⟨100⟩ loops are slightly underestimated in their number, even though the shape of the predicted size distribution matches well with the true distribution.

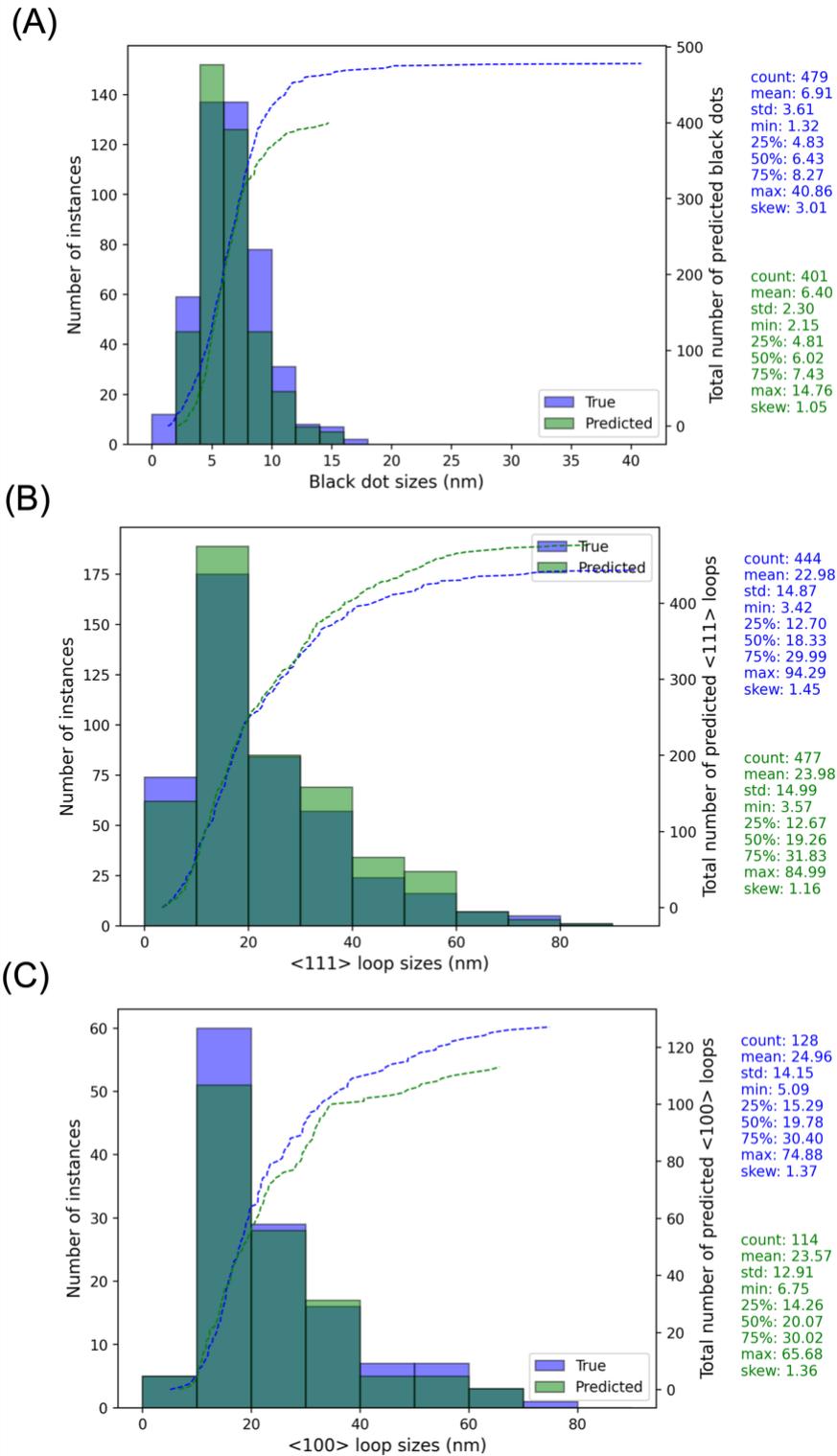

**Figure 5.** Histograms of defect size distributions for all found defects split out by defect type: (A) black dot defects, (B) ⟨111⟩ loop defects and (C) ⟨100⟩ loop defects.

Another useful way to represent the comparisons of true and predicted defect statistics is by way of parity plots. In **Figure 6**, we present parity plots of the true vs. predicted defect shape, size and densities split out by defect type. Each data point plotted in **Figure 6** represents the calculated defect statistics from an individual test image. This analysis is useful for picking out particular images that may perform better or worse than others, as well as identifying problematic outlier images. For example, this analysis enabled us to pick out a single test image with very large number of true black dot defects whose count was severely underestimated by the model (lower right corner in **Figure 6E**). This single test image thus contributed to most of the observed error for the black dot defect densities. While there is some variation in how well individual images are predicted, the model does quite well on the scale of individual images, with mean absolute error values of the per-image defect size of about 3 nm and per-image defect density of about $0.5 \times 10^4$ #/nm$^2$. It is also notable that when taken as an average over the entire test image set, the model predictions improve and become excellent for all three properties of interest. We note here that instead of representing the defect size as nm, one could also assess the error using units of pixels. In addition, instead of assessing defect densities as number of defects per square nm, one could examine the errors in defect counts by counting the total true and predicted defects of each type for each image. We have also examined the errors in the model performance for this dataset using pixels and total defects per image as an assessment of defect size and defect density, respectively (see **Supporting Information**).

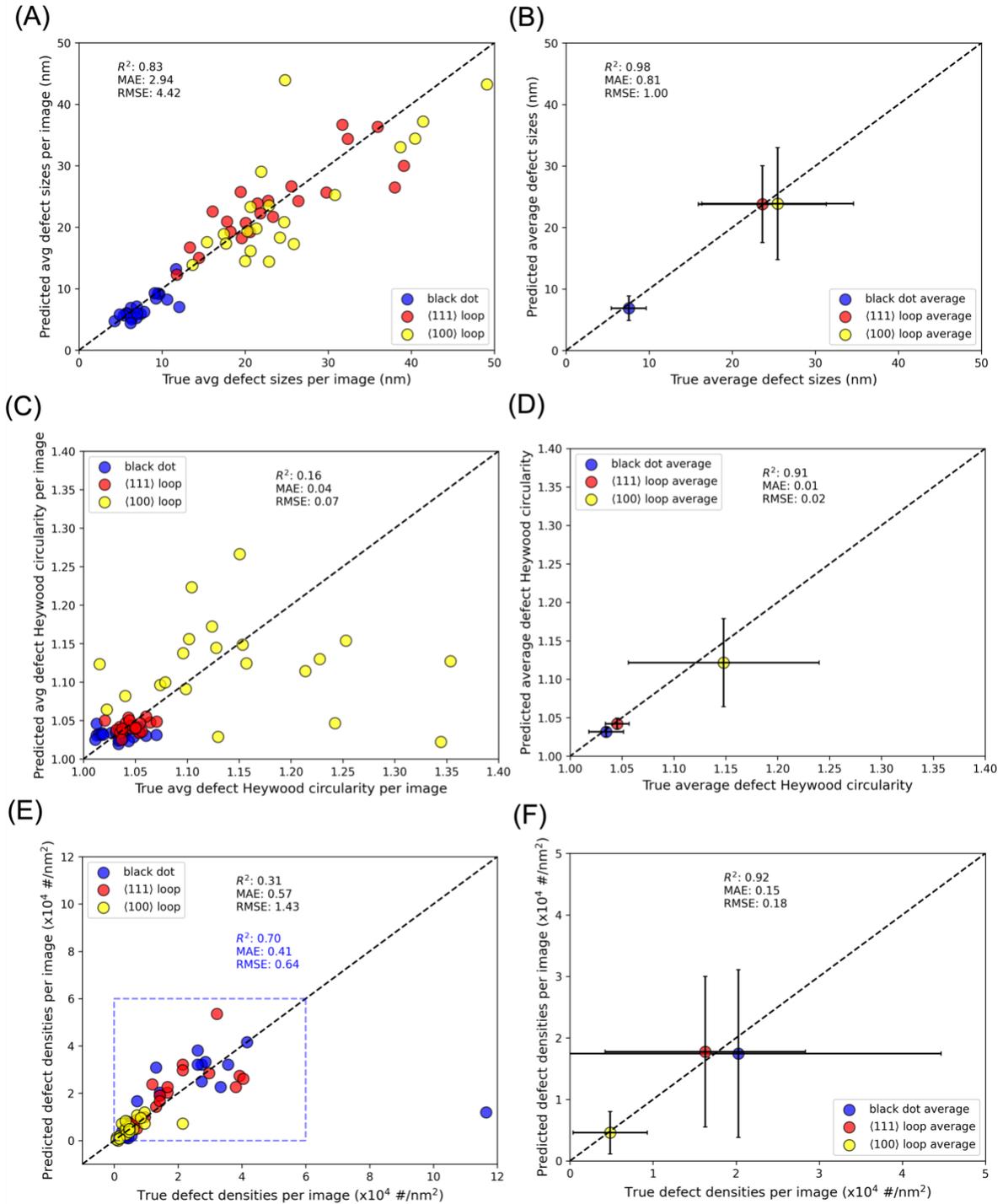

**Figure 6.** Parity plots comparing true and predicted defect sizes (A, B), shapes (C, D), and densities (E, F) on a per-validation image basis (A, C, E, left column) and averaged over all validation images (B, D, F, right column). In all panels, blue, red and yellow points represent values for black dots, ⟨111⟩ loops, and ⟨100⟩ loops, respectively. For the panels averaged over all validation images (B, D, F), the points denote the average value for the respective defect type and the error bars are the standard deviations in the true and predicted values. In (E), the statistics listed

in blue correspond to the datapoints enclosed in the dashed blue box, which removes the single outlier image with significantly underestimated number of black dot defects.

As a final visualization to help further quantify and better understand per-image and overall model errors, we have taken the same per-image data from above and re-cast the values in terms of percent error for each defect type. An example of this result is given in **Figure 7** for the case of defect size errors. Analogous plots of defect shape and defect density errors can be found in the **Supporting Information. Figure 7** enables further comparison between per-image and overall expected errors. For instance, in **Figure 7** it is evident that the defect size percent errors are typically about 30% or lower, and that a single test image shows particularly poor prediction of ⟨100⟩ loop sizes. Further examination of predictions made on this poorly predicted test image show that this large percentage error isn't due to the model predicting many ⟨100⟩ loops poorly in terms of their size, but rather that the model predicts one large loop in particular as ⟨100⟩ when the ground truth indicates it is a ⟨111⟩ loop. This loop is much larger than the other ⟨100⟩ loops in the image, resulting in a large size error. It is also worth noting that, when taken as an average, the per-image errors for defect sizes are under 20%. Further, if the entire distribution of defect sizes is taken together and not separated on a per-image basis, the average errors drop further and are consistently under 10%.

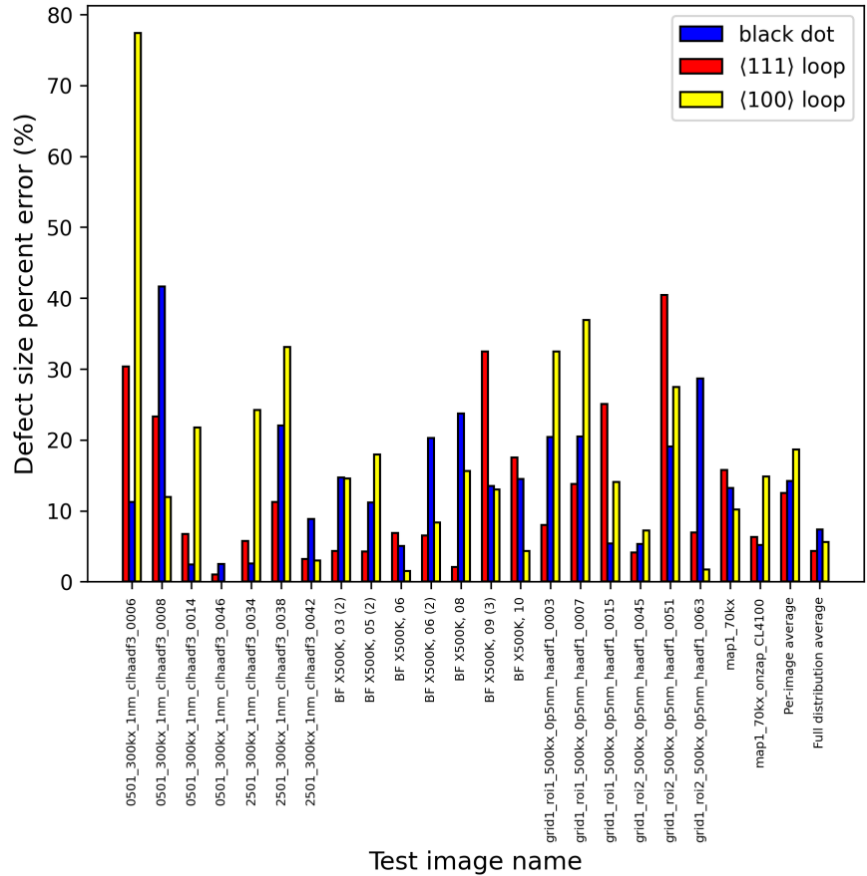

**Figure 7.** Bar plot showing the per-image predicted defect size percent error for each defect type. Also provided on the right-hand side of the plot are the per-image average and the values obtained from the full distribution. The test images shown here are from Dataset1 "initial split".

A major reason for quantifying the defect type, size, shape and density is that these properties play a role in determining alloy mechanical properties. As mentioned in the introduction, the dispersed barrier hardening model uses information of defect type, size and number density to determine the increase in material yield or ultimate tensile strength (hardening) resulting from the creation of defects. Typically, only average size and density information is readily available. However, with the use of the present data and models, the full size distributions and more detailed defect density data for each defect type are available, enabling a more detailed analysis of hardening. Here, we compare the machine learning predicted radiation induced hardening for the present data to the hand counting ground truth value. Following the work of Field et al.,[33] we use the simplified dispersed barrier hardening model with materials constants from Field et al.[33] (see **Supporting Information** for more details), and calculate the expected (from the ground truth) and predicted (from the Mask R-CNN predictions on test images) hardening. In practice, this is done by calculating the hardening contribution of each defect type for each image, then summing the contributions together to obtain the total hardening. This summing step can be done by either simply adding all the contributions (linear sum) or adding the squares of the contributions and taking the square root of this sum of squares (quadrature sum), and it is often unclear which method is best when mixed features are present in the microstructure, so we have done both here.[33] From this analysis, we find that, depending on the image examined, the hardening amount ranges from about 50-200 MPa. Further, we find that the mean absolute error between true and predicted hardening is 16.05 (11.05) MPa based on linear (quadrature) sum, respectively. These absolute error values translate into mean absolute percent errors of 12.9% (13.7%) for linear (quadrature) sum, respectively. These findings indicate that the present Mask R-CNN model predictions of defect sizes and densities can be used to predict the expected hardening with an average error in the range of 10-20 MPa, which is approximately 10% of the total expected hardening based on the observable defects in the images. Other, non-observable features, such as small vacancy and interstitial clusters that exist under the resolution limit of the TEM used as well as precipitates are not considered.

**Understanding variations in model performance based on training and testing data choice: random cross validation**

The performance of machine learning models of all types can be sensitive to the choice of data sets used for training and testing. In object detection, cross validation is not typically performed, as the data set sizes for both training and testing are often very large (e.g. a few million instances). In the limit of large datasets, cross validation will typically not yield significantly different results in the model predictions, as the training and test sets are sampled from the same domain, and cross validation can become computationally impractical. However, for more specific object detection applications such as the present work of finding defects in irradiated alloys, the volume of data is typically much smaller, often on the order of a few thousand instances instead of a few million.

Here, to assess the sensitivity of model performance to the choice of which images are used for training and testing, we perform random cross validation of the train and test sets. This process consists of making five random splits of the images, always holding 21 images out for testing and using the remaining images for training. Splitting the images in this way makes it so about 15-20% of the total defects are reserved for testing, and that the training and testing sets are drawn from roughly the same domain

**Table 5** summarizes the results for the random leave out cross validation test. From **Table 5**, it is evident that the effect of different images used in training and for testing is moderate in scale, with ranges (standard deviations) of the overall defect find F1 score, overall defect type F1 score, average defect size error (all defects) and average defect density error of 0.04 (0.02), 0.05 (0.02), 9.25% (3.80%), and 13.65% (5.31%), respectively. These ranges and standard deviations in key statistics are larger than what was found from running the same model multiple times to assess model randomness (see **Supporting Information**), which indicates that the choice of training and test images, at least for this particular application, may yield meaningfully different predictions of model performance.

From **Table 5**, we can see that Split 2 results in the best density predictions but worst defect size predictions of the set of five splits conducted here, while Split 4 results in the worst density predictions but best defect size predictions. **Figure 8** provides parity plots visualizing these best and worst cross validation splits for predicting defect size and defect density. An observation from **Figure 8** is that the error values between best and worst cross validation split differ by factors ranging from about 1.5-2.5. More specifically, the RMSE of defect density changing from 0.70 $\times 10^4$ #/nm$^2$ (best) to 1.74 $\times 10^4$ #/nm$^2$ (worst) is a factor of 2.5 and RMSE of defect size changing from 6.00 nm (best) to 8.83 nm (worst) is a factor of 1.5. For the defect size error, one test image is the main culprit for the worsened trend, which can be traced to poor predictions of ⟨100⟩ loop defect sizes for one image. We speculate this error is due to missing instances of ⟨100⟩ loops and misidentifying other defect types as ⟨100⟩ loops, thus pushing the average ⟨100⟩ loop size for this image to a small value. For the defect density error, three test images showed significant underprediction of defects, which for all cases were instances of the model significantly underestimating the number of black dot defects. Overall, this analysis indicates that, just as in the case of non-deep learning machine learning applications, performing numerous splits of cross validation is useful for obtaining a more informed assessment of the model performance.

**Table 5.** Summary of random leave out cross-validation test results. Models were trained for 300,000 iterations for all runs.

|  | Split 1 | Split 2 | Split 3 | Split 4 | Split 5 | Average | Range | Standard deviation |
|---|---|---|---|---|---|---|---|---|
| **Number of tests instances** | 1051 | 892 | 1136 | 1284 | 972 | 1067 | 392 | 135.5 |
| **Overall F1 score @ IoU=0.3** | 0.81 | 0.80 | 0.80 | 0.77 | 0.80 | 0.80 | 0.04 | 0.02 |
| **Overall Defect Type F1 score @ IoU=0.3** | ⟨100⟩: 0.67<br>⟨111⟩: 0.82<br>bdot: 0.83<br>Overall: 0.77 | ⟨100⟩: 0.73<br>⟨111⟩: 0.86<br>bdot: 0.88<br>Overall: 0.82 | ⟨100⟩: 0.66<br>⟨111⟩: 0.84<br>bdot: 0.84<br>Overall: 0.78 | ⟨100⟩: 0.69<br>⟨111⟩: 0.82<br>bdot: 0.82<br>Overall: 0.78 | ⟨100⟩: 0.74<br>⟨111⟩: 0.83<br>bdot: 0.86<br>Overall: 0.81 | 0.79 | 0.05 | 0.02 |
| **Defect size (all defects) (% error)** | ⟨100⟩: 5.59<br>⟨111⟩: 4.36<br>bdot: 7.40<br>Overall: 5.78 | ⟨100⟩: 15.73<br>⟨111⟩: 6.09<br>bdot: 12.44<br>Overall: 11.42 | ⟨100⟩: 14.98<br>⟨111⟩: 4.93<br>bdot: 12.18<br>Overall: 10.70 | ⟨100⟩ 2.56<br>⟨111⟩: 1.94<br>bdot: 2.01<br>Overall: 2.17 | ⟨100⟩: 4.41<br>⟨111⟩: 7.61<br>bdot: 7.35<br>Overall: 6.46 | 7.31 | 9.25 | 3.80 |
| **Defect size (found defects) (% error)** | ⟨100⟩: 3.95<br>⟨111⟩: 0.21<br>bdot: 7.45<br>Overall: 3.87 | ⟨100⟩: 6.85<br>⟨111⟩: 1.11<br>bdot: 13.85<br>Overall: 7.27 | ⟨100⟩: 14.61<br>⟨111⟩: 6.75<br>bdot: 3.28<br>Overall: 8.21 | ⟨100⟩: 0.95<br>⟨111⟩: 0.45<br>bdot: 12.34<br>Overall: 4.58 | ⟨100⟩: 1.23<br>⟨111⟩: 5.88<br>bdot: 2.40<br>Overall: 3.17 | 5.42 | 5.04 | 2.20 |
| **Defect density (% error)** | ⟨100⟩: 10.94<br>⟨111⟩: 7.43<br>bdot: 16.28<br>Overall: 11.55 | ⟨100⟩: 9.26<br>⟨111⟩: 14.21<br>bdot: 1.20<br>Overall: 8.22 | ⟨100⟩: 10.07<br>⟨111⟩: 6.90<br>bdot: 13.99<br>Overall: 10.32 | ⟨100⟩: 12.27<br>⟨111⟩: 22.77<br>bdot: 30.56<br>Overall: 21.87 | ⟨100⟩: 0.00<br>⟨111⟩: 0.70<br>bdot: 33.72<br>Overall: 11.48 | 12.69 | 13.65 | 5.31 |

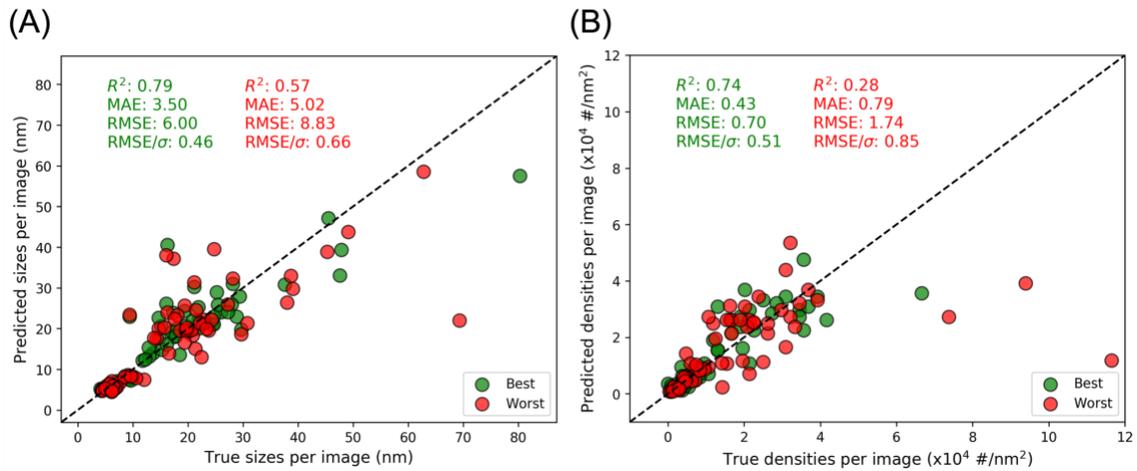

**Figure 8.** Parity plots showing the predicted vs. true defect sizes (A) and densities (B), where each data point results from a specific test image. The green and red data denote the best and worst CV split for each quantity, respectively.

**Understanding limitations of model performance and domain based on training and testing data choice: targeted grouped cross validation**

In addition to random leave out tests, it has been demonstrated in other machine learning applications of materials science that leaving out physically-motivated groups of data is a useful method to more selectively probe model performance.[43–45] Therefore, our second cross validation test consists of leaving out physically motivated groups of images in an attempt to more rigorously evaluate the domain of applicability of our model. These leave out group (LOG) tests are described as follows: LOG Test 1 (leave out irradiation condition): This test keeps the alloys consistent between train and test image sets, but the irradiation conditions between the train and test sets are different. These irradiation conditions differences make it such that the training set will be on smaller ⟨111⟩ loops and ⟨100⟩ loops and a higher density of black dots on two alloys compared to the larger loops and lower density of black dots in the test set. LOG Test 2 (leave out alloy test): This test keeps the irradiation conditions consistent between train and test image sets, but the alloys are different. These composition and sink density differences make it such that the training set will have large loops compared to the test set. LOG Test 3 (leave out sample and microscope type): This test keeps groups in the domain based on the microscope and sample used. The training dataset images were acquired on an older microscope (Philips CM200) with simple starting microstructures while the test dataset images were acquired on a newer microscopes (FEI Talos F200X or JEOL 2100F) with samples that have a more complicated microstructure. The training dataset was obtained entirely by Kevin Field, while the test dataset has two microscopists one of whom was Kevin Field while the other was Dalong Zhang. [46]

**Table 6** summarizes the results for the leave out group cross validation tests. From **Table 6**, a few key results emerge. First, the overall defect type F1 scores for the leave out group tests are generally lower, in the range of 0.55-0.69, than the overall defect type F1 scores obtained from the random leave out cross validation tests, which were in the range of 0.77-0.82. Both the lower values of the overall defect type F1 scores and their larger range for the leave out group tests vs. the random leave out tests make sense. The F1 scores are lower for leave out group tests because it is a more demanding test of the model, as the test images are further outside the domain of the training data than for the random cross validation test, where the training and test data are drawn from the same domain of images. As the training and test image sets are more similar for each iteration of random cross validation, the range of reported F1 scores is smaller. The leave out group tests examined here contain different train/test splits which differ markedly in their character, resulting in a larger range of model performance quality.

In addition to differences of model performance between random vs. leave out group cross validation tests, we can assess the change in model performance for the leave out group test when the training dataset for each test is changed from using the initial Dataset2 to the newer Dataset2 expanded dataset. The performance differences in the leave out group tests between the use of Dataset2 vs. Dataset2 expanded for training suggests that, for these more demanding tests, the larger amount of training data contained in Dataset2 expanded is useful from the standpoint of broadening the domain of applicability of the model. For example, for the leave out alloy test, the overall defect type F1 score increased from 0.55 to 0.64 when training on Dataset2 vs. Dataset2 expanded, and for the leave out microscope/sample test, the overall defect type F1 score increased from 0.60 to 0.69. For the leave out irradiation test, the F1 score remained approximately unchanged between training for the two different datasets. By inspecting the defect type F1 score per defect type, we can see that the improvement in model performance for the leave out alloy and leave out microscope/sample tests is due to different factors. For the leave out alloy test, the improvement in defect type F1 stems from improvements in F1 scores of all three defect types. In contrast, for the leave out microscope/sample test, the improvement in defect type F1 comes from improvement in correctly identifying the ⟨100⟩ loops only.

**Table 6.** Summary of leave out group cross validation test results.

| Group test | Dataset type | Number of train images (defects), number of defects per type | Number of test images (defects) | Defect ID F1 @ IoU = 0.3 | Defect find F1 @ IoU = 0.3 |
|---|---|---|---|---|---|
| Leave out irradiation | Dataset2 | 12 (370) bdot: 117 ⟨111⟩: 195 ⟨100⟩: 58 | 9 (649) | bdot: 0.86 ⟨111⟩: 0.85 ⟨100⟩: 0.26 Overall: 0.66 | 0.79 |
| Leave out irradiation | Dataset2 expanded | 21 (1340) bdot: 707 | 9 (649) | bdot: 0.85 ⟨111⟩: 0.81 | 0.80 |

| | | ⟨111⟩: 423<br>⟨100⟩: 210 | | ⟨100⟩: 0.22<br>Overall: 0.63 | |
| Leave out alloy | Dataset2 | 9 (649)<br>bdot: 268<br>⟨111⟩: 334<br>⟨100⟩: 47 | 51 (6837) | bdot: 0.80<br>⟨111⟩: 0.50<br>⟨100⟩: 0.36<br>Overall: 0.55 | 0.69 |
| Leave out alloy | Dataset2 expanded | 18 (1732)<br>bdot: 767<br>⟨111⟩: 651<br>⟨100⟩: 314 | 51 (6837) | bdot: 0.87<br>⟨111⟩: 0.62<br>⟨100⟩: 0.43<br>Overall: 0.64 | 0.66 |
| Leave out microscope/sample | Dataset2 | 18 (1606)<br>bdot: 598<br>⟨111⟩: 792<br>⟨100⟩: 216 | 70 (3285) | bdot: 0.81<br>⟨111⟩: 0.68<br>⟨100⟩: 0.33<br>Overall: 0.60 | 0.75 |
| Leave out microscope/sample | Dataset2 expanded | 69 (8569)<br>bdot: 4038<br>⟨111⟩: 2493<br>⟨100⟩: 2038 | 70 (3285) | bdot: 0.82<br>⟨111⟩: 0.68<br>⟨100⟩: 0.57<br>Overall: 0.69 | 0.75 |

**Examining impact of ground truth labeling by domain experts on model performance**

As discussed in the introduction, one issue with characterizing and quantifying defects in electron microscopy images is that the establishment of the ground truth labels is done manually by human domain-expert labelers. This labeling process inherently carries some level of subjectivity with it, as different human labelers may disagree about whether a feature in an image constitutes a defect being present, and the type of defect. In addition, some labelers may exhibit labeling patterns notably distinct from other labelers. For example, in the work of Li et al., when comparing the results of five human labelers quantifying the number and size of defects in a set of images, two labelers differed in their labeling systematically, with one labeler tending to categorize many more image features as defects compared to the other labeler.[28]

Here, we assess the performance of Mask R-CNN models trained on different ground truth datasets. **Table 7** summarizes the key results of this test. Overall, results of both datasets show very similar levels of average accuracy for all test statistics, where the differences in scores between the two datasets is of the same magnitude as observed from our test assessing model randomness (see **Supporting Information**). One notable difference is the Dataset1 model tends to show higher density errors for black dots, and the Dataset2 model tends to show higher size errors for black dots. It is not clear what the cause of these differences is, but we speculate it may relate to the nature of the ground truth labels, where Dataset1 contains many instances of image features labeled as black dot defects that were not labeled as a defect at all in Dataset2. In sum, the Mask R-CNN models trained using different ground truth labels perform very similarly, indicating that, at least for this case, the labeling performed by a particular domain expert may not hold obvious advantages compared to another expert. However, it is worth noting here that if certain biases exist in the ground truth labels, for example a labeler who systematically labels certain ambiguous image features as being black dot defects, this bias will likely translate to the trained model. Since the predictive ability of a model can, as an upper bound, only become as accurate as the ground truth data it is trained on, future work should be devoted to establishing publicly available curated datasets which can be labeled and analyzed by many researchers in the field. This process will then involve subsequent model re-training to converge on the most accurate and predictive model of the most relevant metrics as agreed upon by the greater community.

**Table 7.** Summary of model performance on datasets labeled by different domain experts. Both models were trained for 300,000 iterations.

|  | **Dataset 1** | **Dataset 2** |
|---|---|---|
| **Defect find F1 @ IoU = 0.3** | 0.81 | 0.82 |
| **Defect ID F1 @ IoU = 0.3** | ⟨100⟩: 0.67<br>⟨111⟩: 0.82<br>bdot: 0.83<br>Overall: 0.77 | ⟨100⟩: 0.63<br>⟨111⟩: 0.82<br>bdot: 0.81<br>Overall: 0.75 |
| **Defect size (all defects) (% error)** | ⟨100⟩: 5.59<br>⟨111⟩: 4.36<br>bdot: 7.40<br>Overall: 5.78 | ⟨100⟩: 4.31<br>⟨111⟩: 0.11<br>bdot: 15.37<br>Overall: 6.60 |
| **Defect size (found defects) (% error)** | ⟨100⟩: 3.95<br>⟨111⟩: 0.21<br>bdot: 7.45<br>Overall: 3.87 | ⟨100⟩: 9.17<br>⟨111⟩: 0.87<br>bdot: 9.14<br>Overall: 6.39 |
| **Defect density (% error)** | ⟨100⟩: 10.94<br>⟨111⟩: 7.43<br>bdot: 16.28<br>Overall: 11.55 | ⟨100⟩: 18.66<br>⟨111⟩: 14.09<br>bdot: 5.06<br>Overall: 12.60 |

**Examining effect of data set size on model performance**

Analyzing the impact of training dataset size on the model performance enables one to identify the amount of training data required for the model performance to saturate. In addition, even if the model performance does not improve beyond a certain amount of training data, it is likely the domain of applicability of the model is expanded, as discussed above in the context of the leave out group cross validation tests. In this section, we assess the model performance as a function of training dataset size in two different ways. First, we use our largest dataset, Dataset2 expanded, to generate multiple splits of different leave out percent cross validation tests, ranging from leave out 10% to leave out 90% of the images as test data. With these leave out percent cross validation tests, we assess the performance of the model using parity plots of predicted vs. true defect sizes and defect densities of all test set images. For the second test, we construct learning curves which plot per-defect type F1 scores as a function of number of defects of each defect type used in the training data. For this second test, to construct the learning curves, data from the previously discussed leave out group tests, the leave out percent tests to be discussed in this section, and additional runs using Dataset1 and random cross validation to construct training sets of varying sizes were used.

For our first assessment of the effect of dataset size using leave out percent cross validation, **Figure 9** presents parity plots of defect sizes and defect densities split out by defect type for five cases of different dataset sizes. The dataset size was modified by performing multiple iterations of leave out percent cross validation, with the leave out fraction consisting of 10%, 25%, 50%, 75%, and 90% of the images. Each leave out amount was performed three times, where each time a different random portion of the data was left out for testing. A handful of findings are evident from **Figure 9**. In general, the model performance generally improves as less data is held out (equivalently, as the amount of training data increases). More specifically, as the leave out fraction becomes larger, the ability of the model to predict defect sizes becomes significantly worse on a per-image basis, with the RMSE increasing from 3.20 nm (average of 3 iterations of leave out 10%) to 6.25 nm (average of 3 iterations of leave out 90%), nearly a factor of two increase. Interestingly, while the model performance worsens when leaving out up to 90% of the images, the predictive performance is still impressively robust in the limit of small amounts of training data. This finding may suggest that object detection models like Mask R-CNN may offer useful insights and predictions on rather sparse datasets containing fewer than 1000 training instances, and this will be discussed in more detail below. Regarding the predictions of defect density with different leave out amounts, the trends when examining all of the data as a function of leave out amount do not show as clear of a trend as the case of defect sizes and the trend might be affected by the presence of a few images with very high black dot defect densities. However, if the analysis is instead focused on the region where the true defect density is less than $10 \times 10^4$ #/nm$^2$ (blue dashed boxes in **Figure 9**) which constitutes the vast majority of the images studied in this work, then the errors in defect density clearly increase from $1.07 \times 10^4$ #/nm$^2$ (leave out 10%) to $1.49 \times 10^4$ #/nm$^2$ (leave out 90%). As has been observed in past studies, increasing the amount of training data generally results in reduced prediction errors,[30] and may also help broaden the applicability domain of the model.

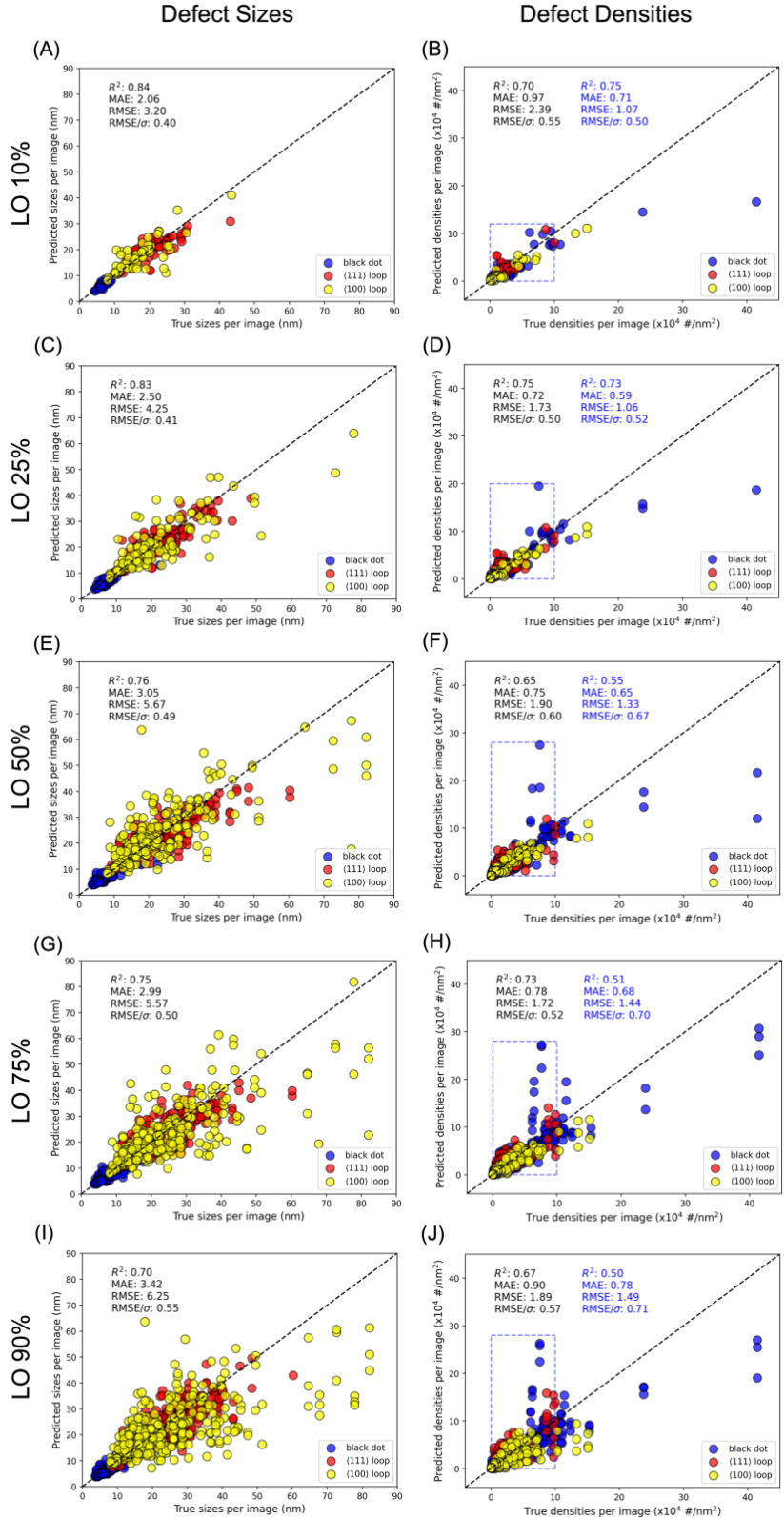

**Figure 9.** Parity plots comparing true and predicted defect sizes (left) and densities (right) for three random splits of 10% (A, B), 25% (C, D), 50% (E, F), 75%, (G, H), and 90% (I, J) cross validation.

The blue, red, and yellow data points denote average values from an individual test image for black dot, $a_0/2\langle 111 \rangle$ and $a_0\langle 100 \rangle$ loops, respectively. For the plots of defect density, the blue dashed box and corresponding statistics are for images where the true densities are less than $10 \times 10^4$ #/nm$^2$.

For our second assessment of the effect of dataset size using all of the cross validation tests described in this work, **Figure 10** contains learning curve plots representing the overall defect type F1 score vs. number of training defects (**Figure 10A**) and the defect type F1 score broken out by defect type vs. number of training defects, this time on a log scale (**Figure 10B**). There are a few key pieces of information we can extract from **Figure 10A.** First, the ability of the model to correctly identify defects quickly increases with number of training defects, with a defect ID F1 score approaching 0.7 for models trained on fewer than 1000 defect instances. After 1000 defects, improvement is incremental with significant diminishing returns, and a defect ID F1 score of about 0.8 is achievable using greater than 6000 defects. Extrapolating these results suggests that achieving a defect ID score meaningfully above 0.8 may require a dataset with greater than 50,000 defects. In **Figure 10A**, the data points for our tests of leave out percent cross validation using Dataset2 expanded (gray triangles) and random leave out cross validation using Dataset1 (gray circles) fall on the same curve. This result makes sense, as both of these methods select training and test images at random. These two datasets differ in the criterion used to select how large the training sets were, and the test image sets used. The random leave out tests (gray circles) used Dataset1, and the test image set was the same in all cases and the number of training images was varied. The leave out percent tests (gray triangles) used Dataset2 expanded, and the test image set changed for each test. The data points corresponding to the leave out group tests (gray squares), except for one instance, always fall below the random cross validation data points for the same amount of training data. This is to be expected, given that the leave out group test is more demanding, and the test data is generally further from the domain of the training data compared to the random cross validation tests.

In **Figure 10B**, we take the same data from **Figure 10A**, but break out the defect ID F1 scores by defect type, and for easier examination of the differences of F1 score between defect types, we plot the number of training defects (i.e. the x-axis) using a log scale. Examining the data in this manner shows that in the limit of very small datasets, e.g., around only 100 defects, the model still performs surprisingly well at correctly identifying black dots and $\langle 111 \rangle$ loops, while there is very poor predictive ability of the $\langle 100 \rangle$ loops. Once the number of black dots and $\langle 111 \rangle$ loops used for training is in the range of a few hundred, the defect ID F1 score is already above 0.8 for these defect types. Thus, expanding the amount of labeled data in our database mainly resulted in the model performing better on the ⟨100⟩ loops, as evidenced by the collection of yellow triangle data points with F1 scores in the range of 0.7-0.75 for the highest defect counts. The increasing trend of ⟨100⟩ loop ID F1 score suggests that the model performance on identifying this defect type still has room for improvement with the inclusion of additional labeled data, even beyond the expanded dataset prepared for this study.

From **Figure 10B**, we can see that the performance of the model in identifying black dots is highest, followed by ⟨111⟩ loops, followed by ⟨100⟩ loops being the worst. This trend is in agreement with the qualitative visual complexity of these defect types: black dots are the most uniform in size, shape and overall appearance and should thus be easiest to categorize, ⟨111⟩ loops are more varied in their size and appearance than black dots but are not as visually diverse as ⟨100⟩ loops, where ⟨100⟩ loops have both edge-on and face-on orientations, yielding a wider range of visually distinct sizes, shapes and contrasts, and the similarity of the edge-on orientation with background line dislocations result in a harder classification task. These qualitative comparisons are also in-line with the leave out group test results, where the black dot predictions between random and leave out group cross validation were effectively identical, while the ⟨111⟩ loop and, in particular, the ⟨100⟩ loop F1 scores were markedly lower for the leave out group tests compared to the random cross validation tests. This performance trend is indicative of black dot defects appearing visually very similar between different groups assigned here, whereas the size, shape and prevalence of the ⟨111⟩ and ⟨100⟩ loops change more dramatically between the train and test sets used for the leave out group tests compared to the random cross validation tests.

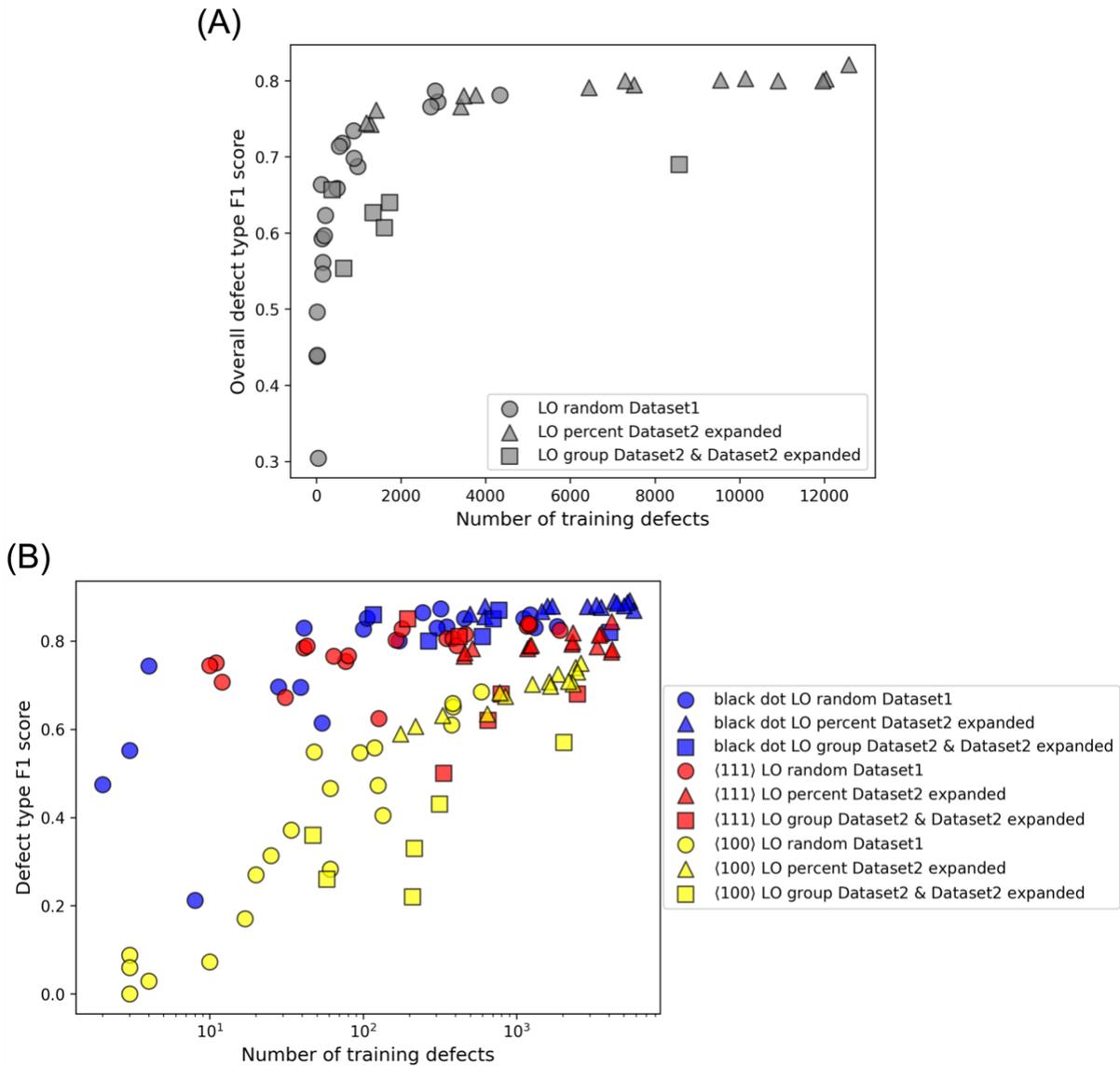

**Figure 10.** Learning curve plots of (A) overall defect type F1 score as a function of number of training defects and (B) defect type F1 score split out by defect type as a function of number of training defects. Note the x-axis of (B) is on a log scale.

## Summary and Outlook

This work and others like it provide an avenue for deep learning models to improve and accelerate materials modeling efforts. Understanding the impact of different irradiation-induced defects in metal alloys on the resulting materials properties and performance hinges on quantifying the numbers, sizes, and shapes of different defect types in the material. The present Mask R-CNN model enables fast, automatic quantification for all of these quantities, as well as refinements to enable more accurate materials modeling by including quantitative data of defect size and shape distributions, instead of just commonly-used average values or models that do not typically include effects related to defect shape.

This work highlights not only the successes and usefulness of deep learning object detection methods for finding defects in microscopy images, but also lays out some of the current limitations and potential issues to be aware of when evaluating the performance of a model. In particular, some high-level findings which may be broadly useful for evaluating model performance can be summarized as follows:

- *Understanding variations in model performance based on data choice*: We have found that the choice of training and test images yield meaningfully different predictions of model performance. As an example, we found the error values for defect size and density errors between best and worst cross validation split differ by factors ranging from about 1.5-2.5. This finding indicates that, just as in traditional machine learning evaluations, cross validation is a useful tool to employ for evaluating performance of object detection models.

- *Understanding limitations of model performance and domain:* The leave out group tests examined in this work contain groups of train and test images which differ markedly in their character, for example, separating sets of images based on alloy type, resulting in a larger range of model performance quality compared to random cross validation. For these more demanding tests, we found that the larger amount of training data contained in our expanded database was useful for broadening the domain of applicability of the model but did not improve the model performance in random cross validation. This finding suggests that expansions of present databases should be focused on including data that exists in different domains from what is already present and that reducing cross validation score may not be a good metric to assess the value of additional data as it misses gains in the domain of applicability of the model.

- *Impact of domain expert labeling to make ground truth:* The generation of ground truth labels can be subjective, leading to different labels from different domain experts. When considering model performance on the same dataset labeled by different experts, we found that very similar levels of average accuracy for all test statistics were obtained, where the differences in scores between the two datasets is of the same magnitude as observed from our test assessing model randomness. However, if certain biases exist in the ground truth labels, for example a labeler systematically labels certain ambiguous image features as being black dot defects, this bias will likely translate to the trained model.

- *Impact of dataset size on model performance:* We found that leaving out up to 90% of the images, the predictive performance is still impressively robust in the limit of small amounts of training data. More specifically, we found that a defect ID F1 score approaching 0.7 for models trained on fewer than 1000 defect instances, while achieving scores significantly above 0.8 was estimated to potentially require more than 50,000 instances. This finding suggests that these models may be reliably trained on datasets that can be generated with modest human labeling efforts of even just a few hours.

We would like to point out that one shortcoming of the present work is that our model is restricted to a single material class (FeCrAl alloys) and uses data for a single STEM imaging condition (bright field, [100] on-zone). Regarding material type, defects like the dislocation loops studied here will manifest with different geometries if the material is changed from, for example, a ferritic steel with the body-centered-cubic crystal structure like the FeCrAl alloys studied here, to an austenitic steel with the face-centered-cubic crystal structure. This change in defect geometry will thus necessitate either training a new model or re-training the present model with these defect instances to increase the model domain and enable accurate predictions on a new material. Regarding imaging condition, analyzing images where the imaging was conducted using a different zone axis (e.g. [111] instead of [100] used here), even for the FeCrAl alloys studied here, will result in the defect loops having different orientations and shapes, e.g. a loop being in plan-view vs. edge-on, and varying image contrasts will change what the model feature map perceives as indicating defected vs. background regions, again necessitating model re-training.

We believe the potential of using object detection models for analyzing electron microscopy images is far from being realized. One area of future work in this space might focus on developing a more general defect model for irradiated alloys that incorporates more than the three defect types considered here, and is further able to classify dislocation lines, cavities and voids formed from gas bubbles, and precipitates, perhaps also taking into account different imaging conditions. Another area of promising future work centers around the exploration and development of methods for synthetic training data generation, including physics-based modeling such as the common "multi-slice" simulations, lower-order models based on simplified assumptions and physical descriptions, and machine learning-centric methods of synthetic data generation such as through the use of generative adversarial networks (GANs).[47] These methods may enable more robust and rapid model training and evaluation, as the reliance on costly and time-consuming experimental data labeling would be reduced, perhaps significantly. A key development to support adoption of these new methods is developing community-based software packages that enables rapid cloud-based dissemination of automated detection packages. To accomplish this, it will be essential to establish a community-agreed on minimum performance metric for the adoption and use of any developed automated defect detection framework. Furthermore, the formation of a robust, community-driven database of labeled TEM images for rapid development and qualification of automated defect detection frameworks will greatly accelerate the development and assessment of new models. Improved data sharing frameworks such as the Materials Data Facility[48] (MDF) and cloud-based services for hosting machine learning models such as DLHub[49,50] are enabling the intersection of materials data and trained machine learning models in a manner that will likely be transformative to the materials research community in the coming years. As a step toward this goal, and in the same spirit as similar efforts of democratization of deep learning models like that of von Chamier et al.,[51] we have made the final trained Mask R-CNN model, images, and analysis scripts publicly available, along with an easy-to-use Google Colab notebook for running the trained model on user-provided images and for re-training the model provided additional labeled data (see **Data Availability**).

The results of the present study demonstrate that the use of standard, off-the-shelf object detection models is extremely effective at quantifying the average size, shape, and density of different object types in the context of defects in electron microscopy images. The findings of this work and findings in recent similar studies[29–32] suggest the maturation of computing hardware (e.g., faster GPUs) and object detection software (e.g., open source Detectron2 package) has reduced the barrier required to perform meaningful object detection tasks. Consistent with these advancements, several companies have developed software packages to aid in performing both traditional computer vision analysis and deep learning analysis of images, including semantic segmentation of objects in images. These tools include Reactiv IP's Smart Image Processing package,[52] Object Research Systems' Dragonfly package,[53] and EPFL's DeepImageJ package,[54] to name a few. Application-specific use of object detection methods with these commercial packages or open source packages like Detectron2, such as model evolution via re-training on newly available data or cloud-based model hosting for broad dissemination, may enable a transformative leap in the manner in which electron microscopy image analysis is performed.

## Acknowledgements

We would like to thank Wisconsin Applied Computing Center (WACC) and Colin Vanden Heuvel for providing access to CPU/GPU cluster, Euler. This work used the Extreme Science and Engineering Discovery Environment (XSEDE), which is supported by National Science Foundation grant number ACI-1548562. Specifically, it used the Bridges-2 system through allocation TG-DMR090023, which is supported by NSF award number ACI-1928147, at the Pittsburgh Supercomputing Center (PSC).[55] Research was sponsored by the Department of Energy (DOE) Office of Nuclear Energy, Advanced Fuel Campaign of the Nuclear Technology Research and Development program (formerly the Fuel Cycle R&D program). Neutron irradiation of FeCrAl alloys at Oak Ridge National Laboratory's High Flux Isotope Reactor user facility was sponsored by the Scientific User Facilities Division, Office of Basic Energy Sciences, DOE. Additional support for K.G.F., D.M. and R.J. was provided by Idaho National Laboratory as part of the Department of Energy (DOE) Office of Nuclear Energy, Nuclear Materials Discovery and Qualification Initiative (NMDQi).

## Author Contributions

R. J. performed the model analysis and wrote the manuscript. M. S. and Y. L. acquired and annotated the data, and performed model analysis. W. H., X. L., R. H., J. G., D. W. Z. X., Z. H. and C. W. annotated the data and performed preliminary model analysis. K. G. F. and D. M. oversaw the project. All authors reviewed the manuscript.

## Data Availability

The datasets generated during and/or analyzed during the current study are available on Figshare (https://doi.org/10.6084/m9.figshare.14691207.v3). The trained model on all images comprising Dataset2 expanded, a Google Colab notebook and associated python scripts to make predictions on new images and save the associated data is also available on Figshare (https://doi.org/10.6084/m9.figshare.14691207.v3). Supporting information discussing the effect of model randomness and model hyperparameters on initial model performance, additional analysis plots of predicted materials properties, and more information the hardening calculations is also available.

**Competing Interests**

The authors declare no competing interests.

# Supporting Information for

Performance, Successes and Limitations of Deep Learning Semantic Segmentation of Multiple Defects in Transmission Electron Micrographs

Ryan Jacobs[1,*], Mingren Shen[1,*], Yuhan Liu[2,*], Wei Hao[2], Xiaoshan Li[2], Ruoyu He[2], Jacob Greaves[1], Donglin Wang[2], Zeming Xie[2], Zitong Huang[2], Chao Wang[2], Kevin G. Field[3], Dane Morgan[12]

[1]Department of Materials Science and Engineering, University of Wisconsin-Madison, Madison, Wisconsin, 53706, USA

[2]Department of Computer Sciences, University of Wisconsin–Madison, Madison, Wisconsin, 53706, USA

[3]Nuclear Engineering and Radiological Sciences, University of Michigan - Ann Arbor, Michigan, 48109 USA

*These authors contributed equally

**Assessing baseline model performance and optimizing model performance on single dataset**

In this section, we assess the out-of-box Mask R-CNN model performance using default Detectron2 settings, optimize model performance via changing hyperparameters, and obtain a baseline model error level from repeated runs from the same initial conditions. All fits in this section are performed on a single dataset, Dataset1. We began this work by training Mask R-CNN models on Dataset1 with varying hyperparameters. In total, Mask R-CNN models with about 25 different hyperparameter configurations were performed. These hyperparameter modifications range from changing the proposed bounding box anchor sizes to the type of neural network used in the backbone (e.g. ResNet 50 vs. ResNet 101) and the origin of the initial starting weights (e.g. COCO vs. ImageNet data).

To compare the performance of the Mask R-CNN models with varying hyperparameters, we focus on the test statistics of overall F1 score, percent error in defect size and percent error in defect density. The values of these test statistics are normalized so they range from 0 (worst result) to 1 (best result). The defect size and density errors are first shifted to be 100-(% error), so that a higher value means better model performance. Since all three metrics are now represented on a unitless scale ranging from 0 to 1, we evaluate the overall model performance by calculating the L2 norm of the three statistics (i.e. the square root of the sum of the squares).

**Figure 11A** contains a sorted bar plot ranking the performance of these 25 models by their L2 norm values. The bars are labeled according to our in-house run labeling scheme, which is only meant to effectively bookkeep which model corresponded to which set of hyperparameters. We have included a summary document (see **Data Availability**) which catalogues which hyperparameters were altered and their values for all of these runs. We can see from **Figure 11** that the model labeled "Run 15" resulted in the best performance (highest total score). The model with these hyperparameters will be used in all other tests in the remainder of this study. We note here that any change to the training data, for example, its size, which images are used in training and validation, or addition of more data in the future may result in a slightly different set of model hyperparameters yielding the best model performance for a particular training configuration. However, we believe such differences in model performance will be small and will not impact the key findings of this work.

The data in **Figure 11** can be used to ascertain how much tuning the model hyperparameters from their default values affected the final model performance, at least according to these three test statistics for this dataset. The model labeled "Run 1" is a run of the Mask R-CNN model from Detectron2 out-of-the-box with all default model parameters and performed well overall compared with the other models evaluated here. To help further illustrate the performance difference attainable with varying hyperparameters, **Figure 11B** contains a radar plot showing the magnitude of the normalized test statistics for the best model (Run 15, blue triangle), the worst model (Run 15b, orange triangle), and a more average performing model (Run 4, green triangle). **Figure 11B** also indicates, at least for this application, that most of the model improvement when changing hyperparameters stems from improvement in the predicted defect densities, followed by a slight improvement in overall F1 score, followed by defect size, which remained largely unchanged for different model hyperparameters.

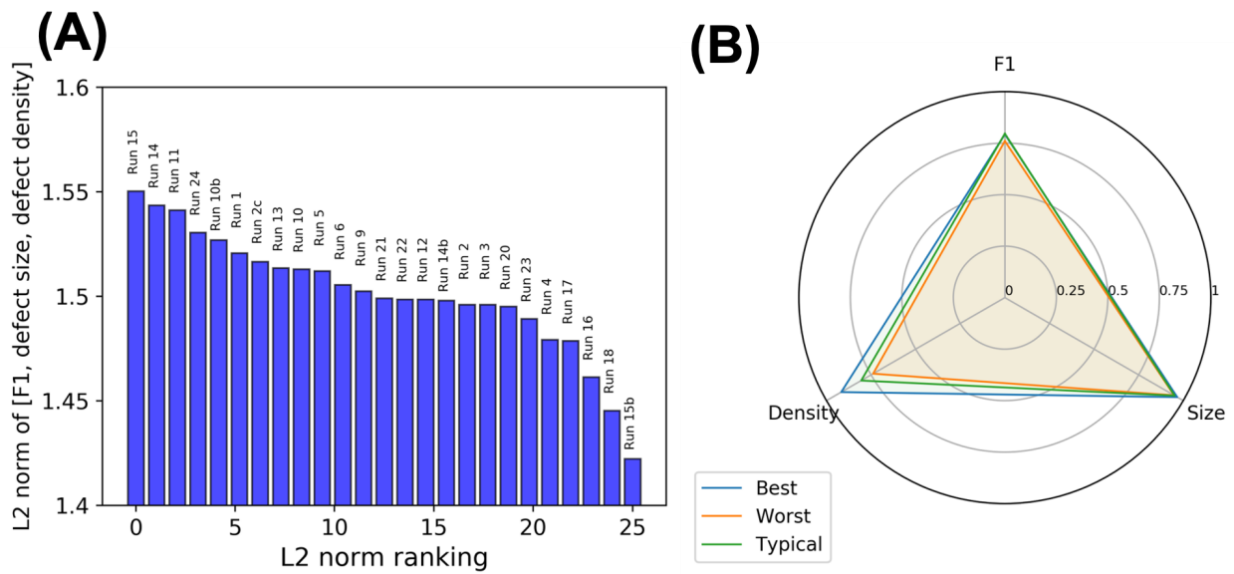

**Figure 11.** (A) Sorted bar chart ranking performance of different Mask R-CNN models according to their calculated L2 norm of the normalized values of overall find F1 score, defect size and defect density. (B) Radar plot showing the performance difference of the best, worst, and a typical model from varying Mask R-CNN model hyperparameters. Note a larger number is associated with better performance. All of these models were evaluated with 100,000 training iterations.

Deep learning neural networks like the Mask R-CNN model used throughout this work are subject to some degree of randomness. This randomness comes from different initial starting weights, dropout layers, random image augmentation (e.g. from chance to flip, rescale, etc.), and different outcomes of the backpropagation algorithm during training, all of which can result in slightly different final model weights. Knowing the magnitude of random error to the test statistics of interest is an important baseline to establish, so one can understand when differences in the reported test statistics are statistically meaningful between different training runs, for example when the training set size is increased or changed to incorporate new images. To test the scale of impact on model randomness on our assessment statistics, we trained the same model three separate times, each starting from the same initial configuration. The summary of the results of this test is provided in **Table 8**. From **Table 8**, it is evident the effect of model randomness is quite minor, on the scale of 0.01 in the overall F1 score, and differences of about 1% (3%) for the average defect size (density) errors, respectively. This model randomness test provides a useful indication of the baseline, smallest expected model error between successive runs, even on the same dataset and model training and evaluation parameters.

**Table 8.** Summary of model randomness test results. Models were trained with 300,000 iterations in all cases.

| Run | 1 | 2 | 3 | Average | Standard deviation |
|---|---|---|---|---|---|
| **Overall F1 score @ IoU=0.3** | 0.81 | 0.80 | 0.80 | 0.80 | 0.01 |
| **Defect size (all defects) (% error)** | ⟨100⟩: 5.59<br>⟨111⟩: 4.36<br>bdot: 7.40<br>Overall: 5.78 | ⟨100⟩: 0.80<br>⟨111⟩: 5.96<br>bdot: 5.60<br>Overall: 4.12 | ⟨100⟩: 7.72<br>⟨111⟩: 2.25<br>bdot: 7.30<br>Overall: 5.76 | 5.22 | 0.95 |
| **Defect size (found defects) (% error)** | ⟨100⟩: 3.95<br>⟨111⟩: 0.21<br>bdot: 7.45<br>Overall: 3.87 | ⟨100⟩: 1.95<br>⟨111⟩: 0.91<br>bdot: 6.01<br>Overall: 2.96 | ⟨100⟩: 0.59<br>⟨111⟩: 1.00<br>bdot: 8.16<br>Overall: 3.25 | 3.36 | 0.46 |
| **Defect density (% error)** | ⟨100⟩: 10.94<br>⟨111⟩: 7.43<br>bdot: 16.28<br>Overall: 11.55 | ⟨100⟩: 2.34<br>⟨111⟩: 7.66<br>bdot: 16.28<br>Overall: 8.76 | ⟨100⟩: 15.63<br>⟨111⟩: 8.11<br>bdot: 19.21<br>Overall: 14.31 | 11.54 | 2.78 |

**Additional analysis plots showing model performance on materials properties**

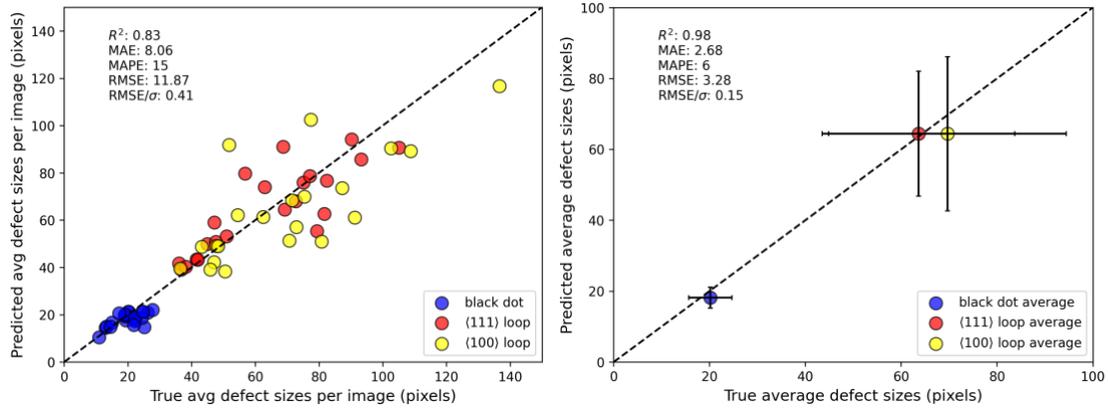

**Figure 12.** (Left) Parity plot of true and predicted defect sizes per image, in units of pixels. (Right) The same data, but represented as average values over all images. The error bars denote standard deviations. These data were from a model fit and evaluated using Dataset1 "initial split".

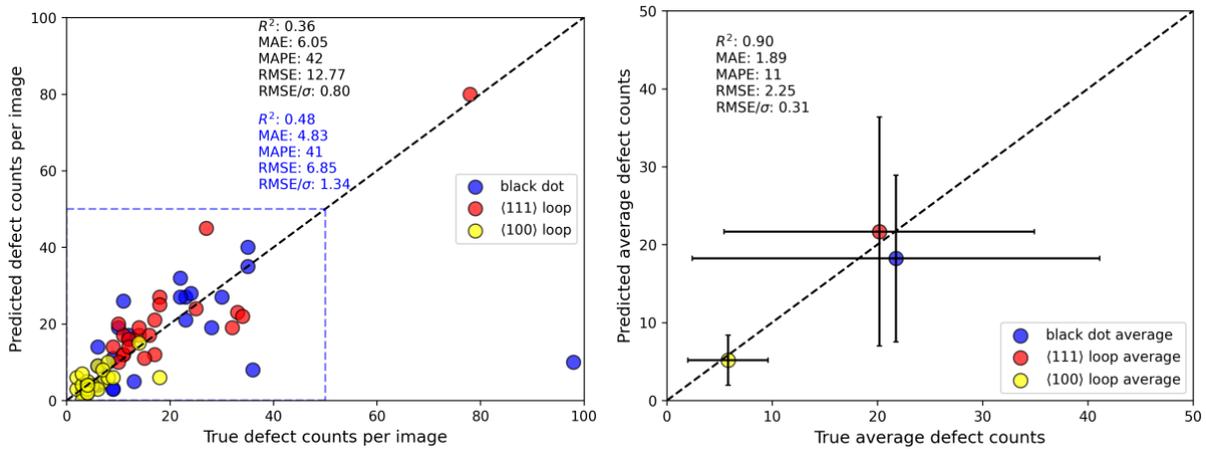

**Figure 13.** (Left) Parity plot of true and defect counts per image. The statistics listed in below were obtained for data points contained in the blue dashed box, denoting images with 50 or fewer defects. (Right) The same data, but represented as average values over all images. The error bars denote standard deviations. These data were from a model fit and evaluated using Dataset1 "initial split".

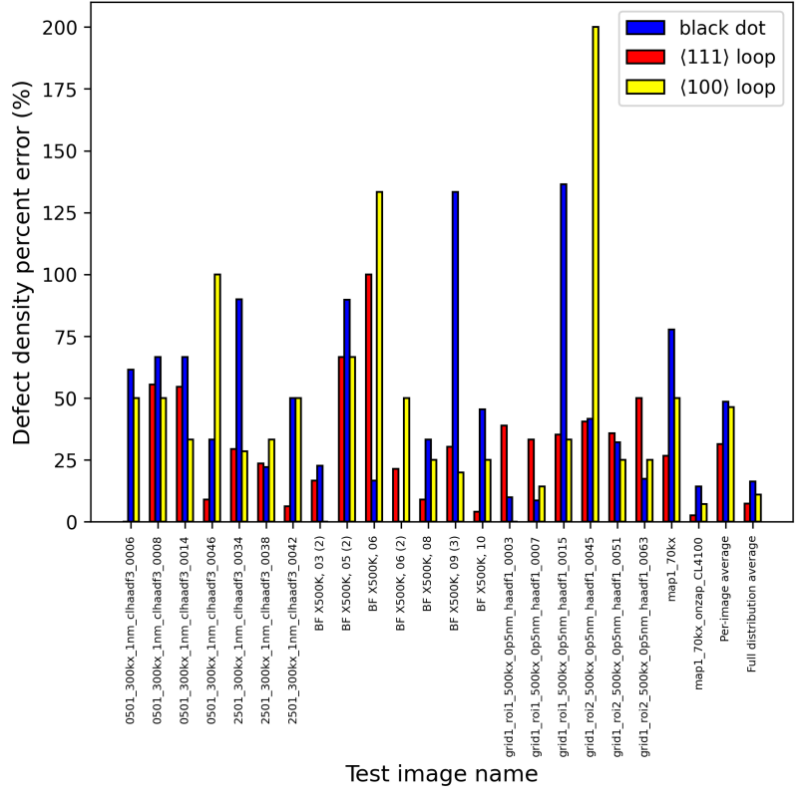

**Figure 14.** Bar plot showing the per-image predicted defect density percent error for each defect type. Also provided on the right-hand side of the plot are the per-image average and the values obtained from the full distribution.

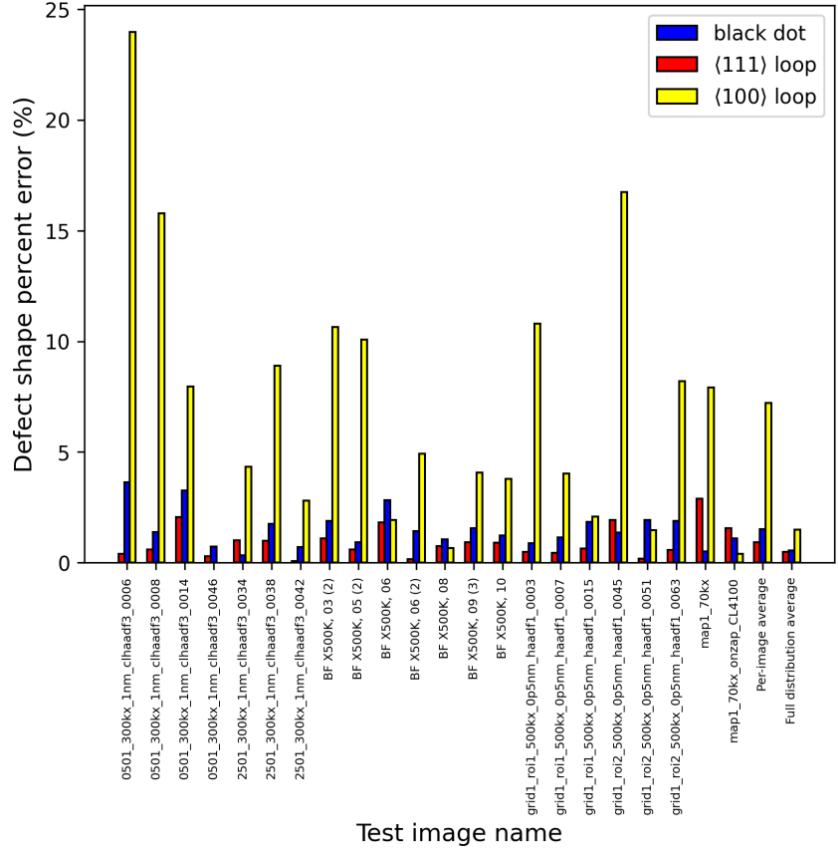

**Figure 15.** Bar plot showing the per-image predicted defect shape percent error for each defect type. Also provided on the right-hand side of the plot are the per-image average and the values obtained from the full distribution.

**Additional information related to calculation of hardening**

The simplified dispersed barrier hardening model is given as (Eq. 1):

$$\Delta\sigma = M\alpha\mu b\sqrt{\rho d} \qquad (1)$$

where Δσ represents the increase in yield or ultimate tensile strength (hardening), *M* is the Taylor value, *α* is the defect barrier strength, *μ* is the shear modulus, *b* is the burgers vector, *ρ* is the defect volume density, and *d* is the defect size distribution. **Table 9** below summarizes the values used for this work.

**Table 9.** Summary of values used for the hardening calculation.

| Variable | Value |
|---|---|
| $M$ | 3.06 (following Refs. 1,2) |
| $\mu$ | 82 GPa |
| $\alpha\_111$ | 0.11 |
| $\alpha\_100$ | 0.32 |
| $\alpha\_bd$ | 0.10 |
| $b$ | 0.249 nm (same for all defects) |

Based on the defect density and size distribution for a given image, the hardening contribution of each defect type is calculated, and the total hardening for a given image can be obtained using either a linear sum (Eq. 2) or a quadratic sum (Eq. 3):

$$\Delta\sigma_{total} = \Delta\sigma_{111} + \Delta\sigma_{100} + \Delta\sigma_{bd} \qquad (2)$$

$$\Delta\sigma_{total} = \sqrt{\Delta\sigma_{111}^2 + \Delta\sigma_{100}^2 + \Delta\sigma_{bd}^2} \qquad (3)$$

To calculate the defect volume densities, the number density of each defect type is needed. The defect areal density is obtained by counting the number of each defect type in an image, and dividing by the image area. The volume is then obtained by this areal density by the thickness of the sample from which the image was taken. **Table 10** below provides a summary of image thickness and pixel/nm conversion data useful for this calculation.

**Table 10.** Summary of image thickness and pixel to nm length conversions.

| Image name | Average thickness (nm) | Pixel to nm conversion |
|---|---|---|
| 2501_300kx_1nm_clhaadf3 | 77.8 | 0.478516 |
| 5401_300kx_1nm_clhaadf3 | 146.5 | 0.478516 |
| mapN_70kx (N= a number) | 87.5 | 0.880523 |
| mapN_70kx_onzap_CL4100 (N= a number) | 111.7 | 0.880523 |
| NROI_100kx_4100CL_foil1 (N= a number) | 113.2 | 0.869141 |
| g1_backonzone | 87.1 | 0.141602 |
| g2_midonzone | 96.3 | 0.141602 |
| 200kV_500kx_p2nm_8cm_CL_grain1 | 100 (not reported) | 0.283203 |
| dalong | 100 (not reported) | 0.283203 |
| grid1_roi1_500kx_0p5nm_haadf1 | 100 (not reported) | 0.283203 |
| BF X500K | 100 (not reported) | 0.283203 |
| 0501_300kx | 100 (not reported) | 0.478516 |
| K713_300kx | 100 (not reported) | 0.478516 |

Finally, the size distribution of each defect type is needed to complete the hardening calculation. To do this, defects are binned into different size ranges, much like the size distribution histograms shown in the main text, e.g., **Figure 4** of the main text. This way, the number of defects corresponding to each size bin for each image is known, e.g., for a given image there may be 5 black dot defects with a binned size value of 5 nm and 3 black dot defects with a binned size value of 10 nm. There is a hardening term corresponding to each of these defect distributions: one contribution for a density of 5 black dots / (image volume) with an assigned size of 5 nm and one contribution for a density of 3 black dots / (image volume) with an assigned size of 10 nm. In this way, the expected hardening for a given image is calculated by summing individual contributions not just from each defect type, but from binned size distributions of each defect type. The total hardening for all images is then obtained using either Eq. 2 or Eq. 3 above.